\documentclass{article} %
\usepackage[preprint]{tmlr}

\usepackage[utf8]{inputenc} %
\usepackage[T1]{fontenc}    %
\usepackage{hyperref}       %
\usepackage{url}            %
\usepackage{booktabs}       %
\usepackage{amsfonts}       %
\usepackage{nicefrac}       %
\usepackage{microtype}      %
\usepackage{xcolor}  
\usepackage{enumitem}

\usepackage[utf8]{inputenc} %

\usepackage{graphicx}

\usepackage[T1]{fontenc}

\IfFileExists{headers/config/showoverfull.config}{
	\overfullrule=1cm
}{
}

\usepackage{marginnote}

\usepackage[backgroundcolor=none,linecolor=red,textsize=footnotesize]{todonotes}

\usepackage{etoolbox}

\newbool{includeappendix}
\setbool{includeappendix}{true} %
\IfFileExists{headers/config/noappendix.config}{
	\setbool{includeappendix}{false}
}{}

\newif\ifincludeappendixx
\ifbool{includeappendix}{
	\includeappendixxtrue
}{
	\includeappendixxfalse
}

\usepackage{xr} %
\usepackage{filecontents}

\ifbool{includeappendix}{}{
	\input{appendix-labels-loader}

	\externaldocument{appendix-labels}
}

\usepackage{xcolor} %

\definecolor{my-full-blue}{HTML}{1F77B4}

\definecolor{my-full-orange}{HTML}{FF7F0E}

\definecolor{my-full-green}{HTML}{2CA02C}

\definecolor{my-full-red}{HTML}{d62728}

\definecolor{my-full-purple}{HTML}{9467bd}

\definecolor{my-full-brown}{HTML}{8c564b}

\definecolor{my-full-pink}{HTML}{e377c2}

\definecolor{my-full-gray}{HTML}{7f7f7f}

\definecolor{my-full-olive}{HTML}{bcbd22}

\definecolor{my-full-cyan}{HTML}{17becf}

\colorlet{my-blue}{my-full-blue!30}
\colorlet{my-orange}{my-full-orange!30}
\colorlet{my-green}{my-full-green!30}
\colorlet{my-red}{my-full-red!30}
\colorlet{my-purple}{my-full-purple!30}
\colorlet{my-brown}{my-full-brown!30}
\colorlet{my-pink}{my-full-pink!30}
\colorlet{my-gray}{my-full-gray!30}
\colorlet{my-olive}{my-full-olive!30}
\colorlet{my-cyan}{my-full-cyan!30}

\definecolor{my-navy-blue}{HTML}{006EB8}

\usepackage{listings}

\usepackage{textcomp}

\usepackage{xcolor}

\usepackage[scaled=0.8]{beramono}

\definecolor{ckeyword}{HTML}{7F0055}
\definecolor{ccomment}{HTML}{3F7F5F}
\definecolor{cstring}{HTML}{2A0099}

\lstdefinestyle{numbers}{
	numbers=left,
	framexleftmargin=20pt,
	numberstyle=\tiny,
	firstnumber=auto,
	numbersep=1em,
	xleftmargin=2em
}

\lstdefinestyle{layout}{
	frame=none,
	captionpos=b,
}

\lstdefinestyle{comment-style}{
	morecomment=[l]//,
	morecomment=[s]{/*}{*/},
	commentstyle={\color{ccomment}\itshape},
}

\lstdefinestyle{string-style}{
	morestring=[b]",%
	morestring=[b]',%
	stringstyle={\color{cstring}},
	showstringspaces=false,%
}

\lstdefinestyle{keyword-style}{
	keywordstyle={\ttfamily\bfseries},
	morekeywords={
		function,
		constructor,
		int,
		bool,
		return,
		returns,
		uint
	},
	morekeywords = [2]{},
	keywordstyle = [2]{\text},
	sensitive=true,
}

\lstdefinestyle{input-encoding}{
	inputencoding=utf8,
	extendedchars=true,
	literate=
	{ℝ}{$\reals$}1%
	{→}{$\rightarrow$}1%
	{α}{$\alpha$}1%
	{β}{$\beta$}1%
	{λ}{$\lambda$}1%
	{θ}{$\theta$}1%
	{ϕ}{$\phi$}1%
}

\lstdefinestyle{escaping}{
	moredelim={**[is][\color{blue}]{\%}{\%}},
	escapechar=|,
	mathescape=true
}

\lstdefinestyle{default-style}{
	basicstyle=\fontencoding{T1}\ttfamily\footnotesize,
	style=numbers,
	style=layout,
	style=comment-style,
	style=string-style,
	style=keyword-style,
	style=input-encoding,
	style=escaping,
	tabsize=2,
	upquote=true
}

\lstdefinelanguage{BASIC}{
	language=C++,
	style=default-style
}[keywords,comments,strings]%

\usepackage{microtype}
\usepackage{graphicx}
\usepackage{booktabs}
\usepackage{threeparttable}
\usepackage{adjustbox}
\usepackage{multirow}
\usepackage{xspace}
\usepackage{wrapfig}
\usepackage{makecell}
\usepackage{dsfont}
\usepackage{subcaption}

\usepackage{hyperref}

\usepackage{amsmath,amsfonts,bm}

\def\eqref#1{equation~\ref{#1}}

\def\1{\bm{1}}

\def\eps{{\epsilon}}

\def\vzero{{\bm{0}}}

\def\vtheta{{\bm{\theta}}}
\def\vlambda{{\bm{\lambda}}}
\def\vsigma{{\bm{\sigma}}}
\def\va{{\bm{a}}}
\def\vb{{\bm{b}}}

\def\vf{{\bm{f}}}

\def\vl{{\bm{l}}}

\def\vu{{\bm{u}}}
\def\vv{{\bm{v}}}

\def\vx{{\bm{x}}}
\def\vy{{\bm{y}}}
\def\vz{{\bm{z}}}

\def\mA{{\bm{A}}}

\def\mI{{\bm{I}}}

\def\mW{{\bm{W}}}

\DeclareMathAlphabet{\mathsfit}{\encodingdefault}{\sfdefault}{m}{sl}
\SetMathAlphabet{\mathsfit}{bold}{\encodingdefault}{\sfdefault}{bx}{n}

\def\gN{{\mathcal{N}}}

\def\sR{{\mathbb{R}}}

\newcommand{\E}{\mathbb{E}}

\newcommand{\R}{\mathbb{R}}

\DeclareMathOperator*{\argmax}{arg\,max}
\DeclareMathOperator*{\argmin}{arg\,min}

\DeclareMathOperator{\sign}{sign}

\usepackage{amsmath}
\usepackage{amssymb}
\usepackage{mathtools}
\usepackage{amsthm,thmtools}
\usepackage{fontawesome5}

\newcommand{\bc}[1]{\mathcal{#1}}
\newcommand{\bs}[1]{\boldsymbol{#1}}
\DeclareMathOperator*{\relu}{ReLU}
\newcommand{\RR}{\relu}

\theoremstyle{plain}
\newtheorem{theorem}{Theorem}[section]
\newtheorem{proposition}[theorem]{Proposition}
\newtheorem{lemma}[theorem]{Lemma}

\theoremstyle{definition}

\theoremstyle{remark}

\newcommand{\emailsbalauca}{\texttt{stefann.balauca@gmail.com}\xspace}

\newcommand{\pgpe}{\textsc{PGPE}\xspace}
\newcommand{\rgs}{\textsc{RGS}\xspace}
\newcommand{\grad}{\textsc{GRAD}\xspace}

\newcommand{\mnbab}{\textsc{MN-BaB}\xspace}

\newcommand{\ibp}{\textsc{IBP}\xspace}
\newcommand{\hboxp}{\textsc{HBox}\xspace}
\newcommand{\crown}{\textsc{CROWN}\xspace}
\newcommand{\crownibp}{\textsc{CROWN-IBP}\xspace}

\newcommand{\deeppoly}{\textsc{DeepPoly}\xspace}

\newcommand{\deepz}{\textsc{DeepZ}\xspace}

\newcommand{\sabr}{\textsc{SABR}\xspace}

\newcommand{\staps}{\textsc{STAPS}\xspace}

\newcommand{\cifar}{CIFAR-10\xspace}
\newcommand{\TIN}{\textsc{TinyImageNet}\xspace}
\newcommand{\mnist}{\textsc{MNIST}\xspace}

\newcommand{\cnnt}{\texttt{CNN3}\xspace}
\newcommand{\cnnty}{\texttt{CNN3-tiny}\xspace}
\newcommand{\cnnf}{\texttt{CNN5}\xspace}
\newcommand{\cnnfl}{\texttt{CNN5-L}\xspace}
\newcommand{\cnns}{\texttt{CNN7}\xspace}

\renewcommand{\th}{\textsuperscript{th}\xspace}

\usepackage{tikz}
\usetikzlibrary{calc,decorations,decorations.pathmorphing}
\usetikzlibrary{positioning,fit,arrows}
\usetikzlibrary{decorations.markings}
\usetikzlibrary{shapes,shapes.geometric}
\usetikzlibrary{shadows,patterns,snakes}
\usetikzlibrary{backgrounds,decorations.pathreplacing,automata}
\usetikzlibrary{intersections}
\usetikzlibrary{angles,quotes}
\usetikzlibrary{plotmarks}
\usetikzlibrary{patterns}
\usetikzlibrary{arrows.meta}
\usetikzlibrary{shapes.misc}
\usetikzlibrary{chains}
\usepackage{siunitx}
\newcolumntype{d}[1]{S[table-format=#1]}
\usetikzlibrary{patterns.meta}

\usepackage[capitalize,noabbrev]{cleveref}

\makeatletter
\newcommand\theHALG@line{\thealgorithm.\arabic{ALG@line}}
\makeatother

\crefformat{section}{\S#2#1#3}

\crefrangeformat{section}{\S#3#1#4\crefrangeconjunction\S#5#2#6}

\crefmultiformat{section}{\S#2#1#3}{\crefpairconjunction\S#2#1#3}{\crefmiddleconjunction\S#2#1#3}{\creflastconjunction\S#2#1#3}

\newcommand{\crefrangeconjunction}{--}

\crefname{listing}{Lst.}{listings}
\crefname{line}{Lin.}{Lin.}
\crefname{appendix}{App.}{App.}

\newcommand{\appref}[1]{%
	\ifbool{includeappendix}{\cref{#1}}{the appendix}%
}
\newcommand{\Appref}[1]{%
	\ifbool{includeappendix}{\cref{#1}}{The appendix}%
}

\title{Gaussian Loss Smoothing Enables Certified Training with Tight Convex Relaxations}

\author{\name Stefan Balauca$^{1\,\text{\faIcon[regular]{envelope}}}$ \quad Mark Niklas Müller$^{2,3}$ \quad Yuhao Mao$^{2}$ \vspace{1mm} \\
 		Maximilian Baader$^{2}$ \quad Marc Fischer$^{4}$ \quad Martin Vechev$^{2}$ \vspace{2mm} \\
      \addr $^{1}$INSAIT, Sofia University ``St. Kliment Ohridski'', Bulgaria \\
	  $^{2}$Department of Computer Science, ETH Zurich, Switzerland \\
	  $^{3}$LogicStar.ai (work done while at ETH Zurich) \\
	  $^{4}$Invariant Labs (work done while at ETH Zurich) \\
	  \faIcon[regular]{envelope}\,Correspondence to \emailsbalauca
}

\def\month{07}  %
\def\year{2025} %
\def\openreview{\url{https://openreview.net/forum?id=lknvxcjuos}} %

\begin{document}

\maketitle

\begin{abstract}
    Training neural networks with high certified accuracy against adversarial examples remains an open challenge despite significant efforts. 
    While certification methods can effectively leverage tight convex relaxations for bound computation, in training, these methods, perhaps surprisingly, can perform worse than looser relaxations. 
    Prior work hypothesized that this phenomenon is caused by the discontinuity, non-smoothness, and perturbation sensitivity of the loss surface induced by tighter relaxations. 
    In this work, we theoretically show that applying Gaussian Loss Smoothing (GLS) on the loss surface can alleviate these issues. 
    We confirm this empirically by instantiating GLS with two variants: a zeroth-order optimization algorithm, called PGPE, which allows training with non-differentiable relaxations, and a first-order optimization algorithm, called RGS, which requires gradients of the relaxation but is much more efficient than PGPE. 
    Extensive experiments show that when combined with tight relaxations, these methods surpass state-of-the-art methods when training on the same network architecture for many settings. 
    Our results clearly demonstrate the promise of Gaussian Loss Smoothing for training certifiably robust neural networks and pave a path towards leveraging tighter relaxations for certified training.
\end{abstract}

\section{Introduction}

The increased deployment of deep learning systems in mission-critical applications has made their provable trustworthiness and robustness against adversarial examples \citep{BiggioCMNSLGR13,SzegedyZSBEGF13} an important topic. As state-of-the-art neural network certification has converged to similar approaches \citep{ZhangWXLLJ22,FerrariMJV22}, increasingly reducing the verification gap, the focus in the field is now shifting to specialized training methods that yield networks with high certified robustness while minimizing the loss of standard accuracy \citep{MullerE0V23,mao2023connecting,palma2024expressive}.

\paragraph{Certified Training}
State-of-the-art (SOTA) certified training methods aim to optimize the network's worst-case loss over an input region defined by an adversarial specification. However, as computing the exact worst-case loss is NP-complete \citep{KatzBDJK17}, they typically utilize convex relaxations to compute over-approximations of this loss \citep{GowalIBP2018,SinghGMPV18,SinghGPV19}. Surprisingly, training methods based on the least precise relaxations (\ibp) empirically yield the best performance \citep{ShiWZYH21}, while tighter relaxations perform progressively worse (left, \cref{fig:intro}). \citet{JovanovicBBV22} and \citep{LeeLPL21} investigated this surprising phenomenon which they call the ``Paradox of Certified Training'', both theoretically and empirically, and found that tighter relaxations induce harder optimization problems. Specifically, they identify the \emph{continuity}, \emph{smoothness}, and \emph{sensitivity} of the loss surface induced by a relaxation as key factors for the success of certified training, beyond its \emph{tightness}. Indeed, \emph{all} state-of-the-art methods are based on the imprecise but continuous, smooth, and insensitive IBP bounds \citep{MullerE0V23,mao2023connecting,palma2024expressive}. However, while these \ibp-based methods improve robustness, they induce severe regularization, significantly limiting the effective capacity and thus standard accuracy \citep{MaoMFV24}. This raises the following fundamental question:
\vspace{-2mm}
\begin{center}
    \emph{Can we enable certified training with tight convex relaxations by addressing the discontinuity, non-smoothness, and perturbation sensitivity, thus obtaining a better robustness-accuracy trade-off?}
\end{center}

\begin{figure}[t]
    \centering
    \resizebox{0.85\linewidth}{!}{
    \begin{tabular}{cccccc}
        \cmidrule[\heavyrulewidth]{1-3}
        \cmidrule[\heavyrulewidth]{5-6} 
        Relaxation & Tightness & \grad [\%] &  & \pgpe [\%] & \rgs [\%] \\
        \cmidrule(rl){1-3} \cmidrule(rl){5-6}
        \ibp & {\colorbox{blue!20!white}{0.55}} & {\colorbox{red!60!white}{91.23}} & \multirow{2}{*}{\Large$\substack{\text{Loss} \\ \text{Smoothing}\\ \Longrightarrow}{}$} & {\colorbox{blue!20!white}{87.02}} & {\colorbox{blue!20!white}{90.46}} \\
        \crownibp & {\colorbox{blue!35!white}{1.68}} & {\colorbox{red!20!white}{88.76}} &  & {\colorbox{blue!35!white}{90.23}} & {\colorbox{blue!35!white}{90.71}} \\
        \deeppoly & {\colorbox{blue!50!white}{2.93}} & {\colorbox{red!40!white}{90.04}} &  & {\colorbox{blue!50!white}{91.53}} & {\colorbox{blue!50!white}{91.88}} \\
        \cmidrule[\heavyrulewidth]{1-3}
        \cmidrule[\heavyrulewidth]{5-6} 
    \end{tabular}
    }
    \vspace{-2mm}
    \caption{Illustration of how Gaussian loss smoothing enables certified training with tight relaxations. We compare the Certified Accuracy [\%] obtained by training a \cnnt network on \mnist $\epsilon=0.1$ with different relaxations using either the standard gradient (\grad, used as baseline) or a gradient estimate computed on the smoothed loss surface (\pgpe and \rgs) with the empirical tightness of the method. Using GRAD methods produces the best results with the least tight \ibp relaxation, which is known as the \textcolor{red}{paradox} of certified training. However, using GLS-based optimization methods \pgpe and \rgs, we obtain a clear \textcolor{blue}{correlation} between relaxation tightness and performance.
    } \label{fig:intro}
    \vspace{-3mm}
\end{figure}

\vspace{-4mm}
\paragraph{This Work: Enabling Certified Training with Tight Convex Relaxations}
In this work we propose a conceptual path forward to overcoming the paradox by addressing the three issues identified by prior works. Our key insight is that the discontinuity, non-smoothness, and perturbation sensitivity of the loss surface can be mitigated by smoothing the training loss with a Gaussian kernel over the network parameter search space. We refer to this approach as \emph{Gaussian Loss Smoothing} (GLS). To instantiate GLS, we propose to augment certified training using two methods: (1) a gradient-free method based on Policy Gradients with Parameter-based Exploration (PGPE) \citep{SehnkeORGPS10} and (2) a gradient-based method based on Randomized Gradient Smoothing (RGS) \citep{starnes2023gaussian}. While both methods approximate GLS which is intractable to compute exactly, they enjoy different benefits: (1) PGPE allows training with non-differentiable relaxations, while (2) RGS is much more efficient than PGPE. We note that the GLS principles, as well as the PGPE and RGS algorithms are independent of the training loss we choose to optimize, and therefore can be applied on top of any certified training method and any convex relaxation. 

Using these GLS methods, we empirically demonstrate that tighter relaxations can indeed lead to strictly better networks, thereby confirming the importance of addressing discontinuity, non-smoothness, and perturbation sensitivity (right, \cref{fig:intro}). Critically, with the more precise \deeppoly relaxation \citep{SinghGPV19}, we show that GLS methods achieve strictly better results than the less precise \ibp. Moreover, we demonstrate that the advantages of GLS improve with increasing network depth, outperforming state-of-the-art methods applied for the same architecture in many settings, particularly when precision matters more. Our results demonstrate the promise of GLS for training certifiably robust neural networks and pave a path towards leveraging tighter relaxations for certified training.

\vspace{-2mm}
\paragraph{Main Contributions}
Our core contributions are: %
\begin{enumerate}[leftmargin=*, itemsep=-3pt, topsep=0pt]
    \item A theoretical investigation showing Gaussian Loss Smoothing (GLS) mitigates discontinuity, non-smoothness, and perturbation sensitivity of the loss surface in certified training with tight relaxations.
    \item A PGPE-based adaptation of certified training that approximates GLS in zeroth-order optimization, enabling training with tight non-differentiable relaxations.
    \item A RGS-based adaptation of certified training that approximates GLS in first-order optimization, requiring differentiable relaxations, but achieving a speedup of up to 40x compared to PGPE.
    \item A comprehensive empirical evaluation of different convex relaxations under GLS with the proposed methods, demonstrating the promise of GLS-based approaches.
\end{enumerate}

\section{Training for Certified Robustness}
\label{sec:related_work}

Below, we first introduce the setting of adversarial robustness before providing a background on (training for) certified robustness. For a detailed notation guide see \cref{sec:notation}.

\subsection{Adversarial Robustness}
We consider a neural network $\vf_{\bs{\theta}}(\vx)\colon \bc{X} \to \R^n$, parameterized by the weights $\bs{\theta}$, that assigns a score to each class $i \in \bc{Y}$ given an input $\vx \in \bc{X}$. This induces the classifier $F \colon \bc{X} \to \bc{Y}$ as $F(\vx) \coloneqq \argmax_i \vf_{\bs{\theta}}(\vx)_i$. We call $F$ locally robust for an input $\vx \in \mathcal{X}$ if it predicts the same class $y \in \mathcal{Y}$ for all inputs in an $\epsilon$-neighborhood $\mathcal{B}^\epsilon_p(x) \coloneqq \{ \vx' \in \mathcal{X} \mid \| \vx - \vx' \|_p \leq \epsilon \}$. To prove that a classifier is locally robust, we thus have to show that $F(\vx') = F(\vx) = y, \forall \vx' \in \mathcal{B}^\epsilon_p(x)$.

\vspace{-1mm}
\paragraph{Adversarial Attacks and Empirical Robustness} Disproving local robustness for a given input $\vx$ is done by finding an \emph{adversarial example} $\vx' \in \mathcal{B}^\epsilon_p(x)$ such that $F(\vx') \neq F(\vx)$. The procedure of searching for adversarial examples is called \emph{adversarial attack}. The most common attack methods \citep{GoodfellowSS14,MadryMSTV18} use first-order gradient information to maximize the loss function associated with $\vx'$. When such a method fails to find an adversarial example, we say that the network is \emph{empirically robust} for the given input $\vx$ and perturbation radius $\epsilon$. 

\vspace{-1mm}
\paragraph{Robustness Guarantees}
Local robustness is equivalent to the log-probability of the target class $y$ being greater than that of all other classes for all relevant inputs, i.e., 
\begin{equation} \label{eq:robustness_guarantee}
	\min_{\vx' \in \mathcal{B}, i \neq y}  f(\vx') _y - f(\vx')_i > 0.
\end{equation}
As solving this neural network verification problem exactly is generally NP-complete \citep{KatzBDJK17}, state-of-the-art neural network verifiers relax it to an efficiently solvable convex optimization problem \citep{BrixMBJL23}. To this end, the non-linear activation functions are replaced with convex relaxations in their input-output space, allowing linear bounds of the following form on their output $f(\vx)$ to be computed:
\begin{equation}
    \mA_{l} \vx + \vb_{l} \leq \vf_{\bs{\theta}}(\vx) \leq \mA_{u} \vx + \vb_{u},
\end{equation}
for some input region $\mathcal{B}^\epsilon_p(x)$. From these symbolic bounds we can obtain concrete numerical bounds, usually in a layer-wise fashion, as $\vl_j = \min_{\vx \in \bc{B}} \mA_{l_j} \vx + \vb_{l_j}$, and $\vu_j$ analogously, for each layer $j$. Hence, for the last layer $m$ we obtain bounds for the score associated with each class $\vl_m \leq \vf(\vx) \leq \vu_m$, which can in turn be used to verify \cref{eq:robustness_guarantee}. %

To obtain (certifiably) robust neural networks, specialized training methods are required.
For a data distribution $(\vx, t) \sim \bc{D}$, standard training optimizes the network parametrization $\bs{\theta}$ to minimize the expected cross-entropy loss 
$\theta_\text{std} = \argmin_\theta \E_\bc{D} [\bc{L}_\text{CE}(\vf_{\bs{\theta}}(\vx),t)]$ with $\bc{L}_\text{CE}(\vy, t) = \ln\big(1 + \sum_{i \neq t} \exp(y_i-y_t)\big)$.
To train for robustness, we, instead, aim to minimize the expected \emph{worst-case loss} for a given robustness specification, leading to a min-max optimization problem:
$
\theta_\text{rob} = \argmin_\theta \mathbb{E}_{\bc{D}} \left[ \max_{\vx' \in \bc{B}^{\epsilon}(\vx) }\bc{L}_\text{CE}(\vf_{\bs{\theta}}(\vx'),t) \right]
$.
As computing the worst-case loss by solving the inner maximization problem is generally intractable, it is commonly under- or over-approximated, yielding adversarial and certified training, respectively.

\vspace{-1mm}
\paragraph{Adversarial Training} optimizes a lower bound on the inner optimization objective. To this end, it first computes concrete examples $\vx'\in \bc{B}^{\epsilon}(\vx)$ that approximately maximize the loss term $\bc{L}_\text{CE}$ and then optimizes the network parameters $\bs{\theta}$ for these examples.
While networks trained this way typically exhibit good empirical robustness, they remain hard to formally certify and are sometimes vulnerable to stronger attacks \citep{TramerCBM20,Croce020a}.

\paragraph{Certified Training} typically optimizes an upper bound on the inner maximization objective. To this end, the robust cross-entropy loss $\bc{L}_\text{CE,rob}(\bc{B}^{\epsilon}(\vx),t) = \bc{L}_\text{CE}(\overline{\vy}^\Delta,t)$ is computed from an upper bound $\overline{\vy}^\Delta$ on the logit differences $\vy^\Delta := \vy - y_t$ obtained via convex relaxations as described above and then plugged into the standard cross-entropy loss.
As this can induce strong over-regularization if the used convex relaxations are imprecise and thereby severely reduce the standard accuracy of the resulting models, current state-of-the-art certified training methods combine these bounds with adversarial training \citep{PalmaIBPR22,MullerE0V23,mao2023connecting,palma2024expressive}.
Throughout this work we will focus on Certified Training and Robustness Guarantees only for the $\ell_\infty$-norm, i.e., $\bc{B}^\epsilon(\vx) = \{\vx' \mid ||\vx - \vx'||_\infty \leq \epsilon\}$, as this is the most common setting in the deterministic certified defences literature.
In the following, we introduce some popular convex relaxations used for neural network verification and training.

\subsection{Convex Relaxations}\label{sec:background_convex}
We now discuss four popular convex relaxations of different precision, investigated in this work.

\begin{figure}[t]
	\centering
	\vspace{-1.5em}
	\subcaptionbox{IBP. \label{fig:ReLU_box}}[.43\linewidth]{
		\scalebox{0.8}{\begin{tikzpicture}
	\draw[->] (-2, 0) -- (2, 0) node[right,scale=0.85] {$v$};
	\draw[->] (0, -1.2) -- (0, 1.2) node[above,scale=0.85] {$y$};

	\def\a{-1}
	\def\b{1}
	\def\al{0.15}
	\coordinate (a) at ({\a},{0});
	\coordinate (b) at ({\b},{\b});
	\coordinate (c) at ({0},{0});
	\coordinate (d) at ({\a},{\b});
	\coordinate (e) at ({\b},{0});
	
	\node[circle, fill=black, minimum size=3pt,inner sep=0pt, outer sep=0pt] at (a) {};
	\node[circle, fill=black, minimum size=3pt,inner sep=0pt, outer sep=0pt] at (b) {};
	\node[circle, fill=black, minimum size=3pt,inner sep=0pt, outer sep=0pt] at (c) {};
	
	\fill[fill=blue!90,opacity=0.3] (a) -- (e) -- (b) -- (d) -- cycle;
	\draw[black,thick] (a) -- (c) -- (b);
	\draw[-] (b) -- ($(b)+(0.3,0.3)$);
	\draw[-] ({\a},-0.4) -- ({\a},0.4);
	\draw[-] ({\b},-0.4) -- ({\b},1.9);
	
	\node[anchor=south west,align=center,scale=0.85] at ({\a},-0.50) {$l$};
	\node[anchor=south west,align=center,scale=0.85] at ({\b},-0.50) {$u$};

\end{tikzpicture}}
    }
	\subcaptionbox{DeepPoly. \label{fig:ReLU_deeppoly}}[.43\linewidth]{
		\scalebox{0.8}{\begin{tikzpicture}
	\draw[->] (-2, 0) -- (2, 0) node[right,scale=0.85] {$v$};
	\draw[->] (0, -1.2) -- (0, 1.2) node[above,scale=0.85] {$y$};

	\def\a{-1}
	\def\b{1}
	\def\al{0.15}
	\coordinate (a) at ({\a},{0});
	\coordinate (a1) at ({\a},{\a});
	\coordinate (b0) at ({\b},{0});
	\coordinate (b) at ({\b},{\b});
	\coordinate (c) at ({0},{0});
	\coordinate (aub) at ({\a},{\b-(\b-\a)*\al});
	\coordinate (bub) at ({\b},{(\b-\a)*\al});
	\coordinate (alb) at ({\a},{\a*\al});
	\coordinate (blb) at ({\b},{\b*\al});
	
	\node[circle, fill=black, minimum size=3pt,inner sep=0pt, outer sep=0pt] at (a) {};
	\node[circle, fill=black, minimum size=3pt,inner sep=0pt, outer sep=0pt] at (b) {};
	\node[circle, fill=black, minimum size=3pt,inner sep=0pt, outer sep=0pt] at (c) {};
	
	\fill[fill=green!90, opacity=0.3] (a) -- (b) -- (b0) -- cycle;
	\fill[fill=blue!90, opacity=0.3] (a1) -- (a) -- (b) -- cycle;
	\draw[black,thick] (a) -- (c) -- (b);
	\draw[-] (b) -- ($(b)+(0.3,0.3)$);
	\draw[-] ({\a},-0.4) -- ({\a},0.4);
	\draw[-] ({\b},-0.4) -- ({\b},1.4);

	\node[anchor=north east,align=center,scale=0.85] at ({\a},-0.20) {$l$};
	\node[anchor=south west,align=center,scale=0.85] at ({\b},-0.50) {$u$};	
	\node[anchor=south east,align=center,scale=0.85] at ({-0.3},0.70) {$y\leq \frac{u}{u-l} (v-l)$};
	\node[anchor=north west,align=center,scale=0.85] at ({0.1},0.00) {$y\geq 0$};
	\node[anchor=north west,align=center,scale=0.85] at (-0.9,-0.8) {$y\geq v$};

\end{tikzpicture}}
    }
	\vspace{-1mm}
	\caption{IBP and \deeppoly relaxations of a ReLU with bounded inputs $v \in [l,u]$. For \deeppoly the lower-bound slope $\lambda$ is chosen to minimize the area between the upper and lower bounds in the input-output space, resulting in the blue or green area.
	} \label{fig:ReLU_transformers}
\end{figure}

\paragraph{IBP}

Interval bound propagation \citep{MirmanGV18, GehrMDTCV18, GowalIBP2018} only considers elementwise, constant bounds of the form $\vl \leq \vv \leq \vu$. Affine layers $\vy = \mW \vv +\vb$ are thus also relaxed as
\begin{equation}
    \tfrac{\mW (\vl+\vu) - |\mW|(\vu-\vl)}{2} + \vb \leq \mW \vv +\vb \leq \tfrac{\mW (\vl+\vu) + |\mW|(\vu-\vl)}{2} +\vb,
\end{equation}
where $|\cdot|$ is the elementwise absolute value. ReLU functions are relaxed by their concrete lower and upper bounds $\RR(\vl) \leq \RR(\vv) \leq \RR(\vu)$, illustrated in \cref{fig:ReLU_box}.

\paragraph{Hybrid Box (HBox)} The HBox relaxation is an instance of Hybrid Zonotope \citep{MirmanGV18} which combines the exact encoding of affine transformations from the \deepz or Zonotope domain \citep{SinghGMPV18,WongK18,WengZCSHDBD18,WangPWYJ18} with the simple IBP relaxation of unstable ReLUs, illustrated in \cref{fig:ReLU_box}. While less precise than \deepz, \hboxp ensures constant instead of linear representation size in the network depth, making its computation much more efficient.

\paragraph{DeepPoly}

\deeppoly, introduced by \citet{SinghGPV19}, is mathematically identical to CROWN \citep{ZhangWCHD18} and based on recursively deriving linear bounds of the form
\begin{equation}
    \mA_l \vx + \va_l \leq \vv \leq \mA_u \vx + \va_u
\end{equation}
on the outputs of every layer. While this handles affine layers exactly, ReLU layers $\vy = \RR(\vv)$ are relaxed neuron-wise, using one of the two relaxations illustrated in \cref{fig:ReLU_deeppoly}:
\begin{equation}
    \boldsymbol{\lambda} \vv \leq \RR(\vv) \leq (\vv-\vl) \frac{\vu}{\vu - \vl},
\end{equation}
where product and division are elementwise.
The lower-bound slope ${\vlambda}  = \mathds{1}_{|\vu| > |\vl|}$ is chosen depending on the input bounds $l$ and $u$ to minimize the area between the upper and lower bounds in the input-output space for each neuron separately. Crucially, a minor change in the input bounds can thus lead to a large change in output bounds when using the \deeppoly relaxation.

\paragraph{CROWN-IBP} To reduce the computational complexity of \deeppoly, \crownibp \citep{ZhangCXGSLBH20} uses the cheaper but less precise \ibp bounds to compute the concrete upper- and lower-bounds $\vu$ and $\vl$ on ReLU inputs required for the \deeppoly relaxation. To compute the final bounds on the network output \deeppoly is used. This reduces the computational complexity from quadratic to linear in the network depth. While \crownibp is not strictly more or less precise than either \ibp or \deeppoly, its precision empirically lies between the two \citep{JovanovicBBV22}.

\paragraph{Relaxation Tightness} While we rarely have strict orders in tightness (only \hboxp is strictly tighter than \ibp), we can empirically compare the tightness of different relaxations given a network to analyze. \citet{JovanovicBBV22} propose to measure the tightness of a relaxation as the AUC score of its certified accuracy over perturbation radius curve. This metric implies the following empirical tightness ordering \ibp $<$ HBox $<$ \crownibp $<$ \deeppoly \citep{JovanovicBBV22}, which agrees well with our intuition.

\vspace{-2mm}
\subsection{The Paradox of Certified Training} \label{sec:paradox}
When training networks for robustness with convex relaxations, higher robustness is achieved by sacrificing standard accuracy. Usually, more precise relaxations induce less overapproximation and thus less regularization, potentially leading to better standard and certified accuracy. However, empirically the least precise relaxation, \ibp, dominates the more precise methods, e.g., \deeppoly, with respect to both certified and standard accuracy (see the left-hand side of \cref{fig:intro}).
This is all the more surprising given that state-of-the-art certified training methods introduce artificial unsoundness into these \ibp bounds to improve tightness at the cost of soundness to reduce regularisation and improve performance \citep{MullerE0V23,mao2023connecting,palma2024expressive}.

\citet{JovanovicBBV22} and \citet{LeeLPL21} explained this paradox, by showing that these more precise relaxations induce loss landscapes suffering from discontinuities, non-smoothness, and perturbation sensitivity (a proxy for difficulty to optimize with gradients), making it extraordinarily challenging for gradient-based optimization methods to find good optima. Thus the key challenge of certified training is to design a robust loss that combines tight bounds with a continuous, smooth, and insensitive loss landscape. In \cref{sec:gausian_loss_smoothing}, we discuss these challenges in more detail and show how to overcome them.

\vspace{-2mm}
\section{Gaussian Loss Smoothing (GLS) for Certified Training} \label{sec:gausian_loss_smoothing}

In this section, we address the optimization issues identified in \cref{sec:paradox}---namely, discontinuity, non-smoothness, and sensitivity of the loss surface---by proposing Gaussian Loss Smoothing (GLS). We first illustrate these issues with toy examples (\cref{open_challenges}), highlighting how smoothing can improve loss landscapes. Then, we formalize this intuition in \cref{sec:theory}, showing that GLS yields continuous, smooth, and less non-convex loss surfaces. Finally, we instantiate GLS with two optimization algorithms, PGPE (\cref{sec:pgpe}) and RGS (\cref{sec:rgs}), which enable effective training with tight (and even non-differentiable) convex relaxations.

\vspace{-2mm}
\subsection{Open Challenges: Discontinuity, Non-smoothness and Sensitivity} \label{open_challenges}

\begin{wrapfigure}{r}{0.46\textwidth}
	\vspace{-5mm}
	\centering
	\subcaptionbox{Discontinuity. \label{subfig:discontinuity}}[.49\linewidth]{
		\includegraphics[width=\linewidth]{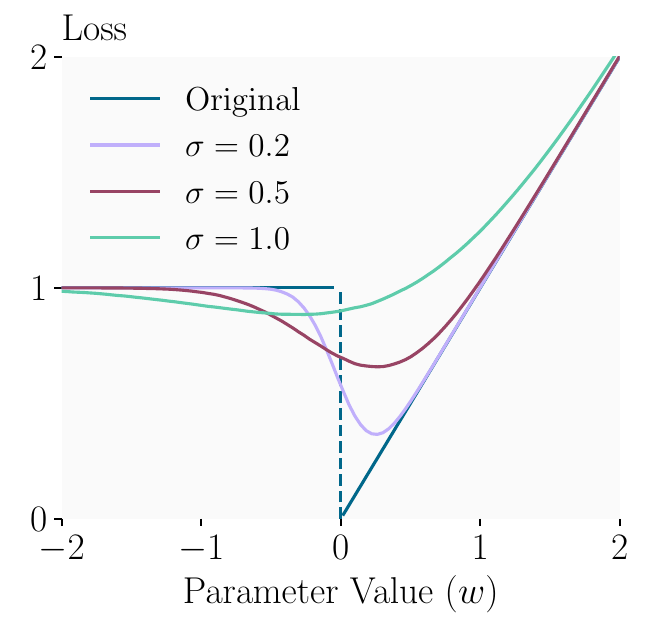}
    }
	\subcaptionbox{Sensitivity. \label{subfig:sensitivity}}[.49\linewidth]{
		\includegraphics[width=\linewidth]{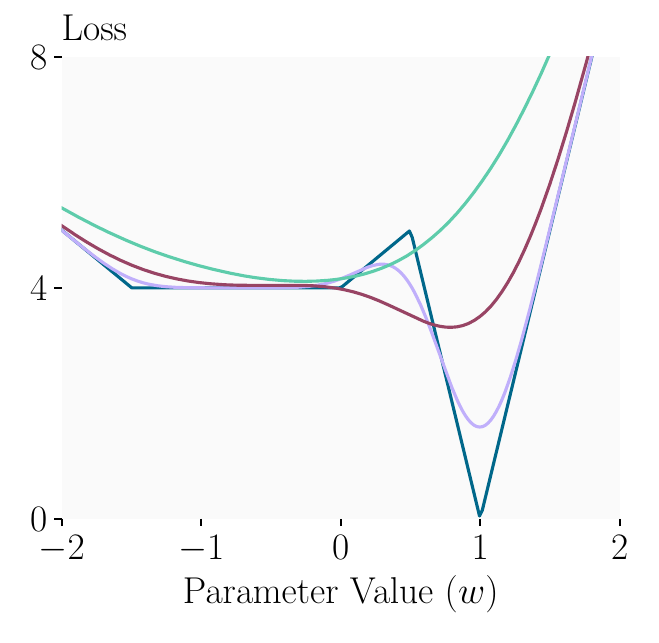}
    }
	\vspace{-2mm}
	\caption{Illustrating the effect of Gaussian Loss Smoothing on the discontinuity (left) and sensitivity of loss functions (right).
	} \label{fig:smoothed_dp}
	\vspace{-6mm}
\end{wrapfigure}

Recall \cref{sec:paradox}, where we discussed the key challenges of certified training with tighter relaxations, namely discontinuity, non-smoothness, and sensitivity of the loss surface.
We now illustrate these key challenges on a toy network and loss in \cref{fig:smoothed_dp}.

On the left-hand side (\textcolor{my-navy-blue}{Original} in \cref{subfig:discontinuity}), we show the \deeppoly lower bound of the one-neuron network $y = \relu(x + w) + 1$ for $x \in [-1, 1]$ over the parameter $w$.
As the original bound $l = 1 + \mathds{1}_{w>0} \cdot (w-1)$ is discontinuous at $w=0$, a gradient-based optimization method initialized at $w>0$ will decrease $w$ until it has moved through the discontinuity and past the local minimum.

The second key factor, non-smoothness, is originally defined as the variation of loss values along the optimization trajectory. For brevity, we restrict this to the \emph{Lipschitz continuity} of the loss function, as a Lipschitz continuous loss function has bounded variation of loss values. A function is called Lipschitz continuous if there exists a constant $L$ such that $|f(\vx) - f(\vy)| \leq L\|\vx-\vy\|$ for all $\vx, \vy$. As \deeppoly has discontinuities, it is not Lipschitz continuous. We remark that Lipschitz continuity is particularly important for gradient-based optimization methods, as this controls the theoretical convergence of such methods.

The third key factor, sensitivity, can be interpreted as the difficulty to optimize with gradients. \citet{JovanovicBBV22} show that \deeppoly is more sensitive than IBP, thus gradient-based optimization methods are more likely to get stuck in bad local minima. We illustrate this with the toy function shown in \cref{subfig:sensitivity}. Here the original function has a bad local minimum for $w \in [-1.5, 0]$ that a gradient-based optimizer can get stuck in. To analyze the badness of a loss surface for gradient-based optimization, we measure the \emph{deviation from convexity} of the loss function, defined to be $D(f) := \max_{\vx, \vy \in \R^d; \lambda \in [0,1]} \delta[f; \vx, \vy, \lambda]$, where $\delta[f; \vx, \vy, \lambda] := f(\lambda \vx + (1-\lambda) \vy) - \lambda f(\vx) - (1-\lambda) f(\vy)$. If a function has a non-positive deviation from convexity, it is convex, thus gradient-based methods can find global optimum. Since this directly measures non-convexity, intuitively, a function with smaller deviation from convexity is easier to optimize with gradient-based methods. We remark that sensitivity as defined in \citet{JovanovicBBV22} is different to the deviation from convexity, but the two are closely related in that both indicate how difficult it is to optimize a function with gradients.

\subsection{Gaussian Loss Smoothing for Certified Training} \label{sec:theory}
\vspace{-1mm}

We now discuss how Gaussian Loss Smoothing can address these challenges.
The central result in this section is formalized in \cref{thm:gls} (proof in \cref{app:proofs}):
\begin{restatable}{theorem}{glsinfdiff} \label{thm:gls}
	Let the parameter $\vtheta \in \sR^d$. Let the nonnegative loss function $L(\vtheta): \sR^d \rightarrow \sR$ have bounded growth, that is, $L(\vtheta) \exp(-\|\vtheta\|^{2-\delta}) \le M$ for some $\delta<2$ and $M>0$. Then, the loss smoothed by an isotropic Gaussian $\gN(\vzero, \sigma^{2} \mI)$, defined as $L_\sigma(\vtheta) := \E_{\bs{\epsilon} \sim \gN(\vzero, \sigma^{2} \mI)} L(\vtheta+\bs{\epsilon})$, is infinitely differentiable. In addition, the deviation from convexity of the smoothed loss never exceed that of the original loss, that is, $D(L_\sigma) \le D(L)$; equality holds iff $L$ is an affine function. Further, assuming $\vtheta$ is in a compact set throughout optimization, $L_\sigma$ is also Lipschitz continuous.

\end{restatable}

\cref{thm:gls} shows several desired qualities of GLS. First, it shows that GLS can turn any discontinuous loss function into a continuous one that is differentiable everywhere, as visualized in \cref{subfig:discontinuity}.
Second, GLS can make the loss surface Lipschitz continuous if we optimize in a compact set, thus ensuring that the loss surface is smooth.
Third, GLS can help to overcome the sensitivity issue since it provably reduces the deviation from convexity as long as the loss function is not affine. As we show in \cref{subfig:sensitivity}, depending on the standard deviation used for smoothing, the local minimum can be reduced or removed, and the loss landscape is thus more favorable. However, the choice of standard deviation is crucial. While a too-small standard deviation only has a minimal effect on loss smoothness and might not remove local minima, a too-large standard deviation can oversmooth the loss, completely removing or misaligning the minima. We again illustrate this in \cref{subfig:sensitivity}. 
There, a small standard deviation of $\sigma=0.5$ works properly, while $\sigma=0.25$ does not smooth out the local minimum, and $\sigma=1.0$ severely misaligns the new global minimum with that of the original function.
Overall, GLS has the theoretical potential to mitigate the key issues, discontinuity, non-smoothness, and sensitivity, for tight convex relaxations (as identified by \citet{JovanovicBBV22} and \citet{LeeLPL21}).

\begin{figure}[t]
    \centering
	\vspace{-4mm}
	\begin{subfigure}[t]{0.24\textwidth}
		\centering
        \includegraphics[width=\linewidth]{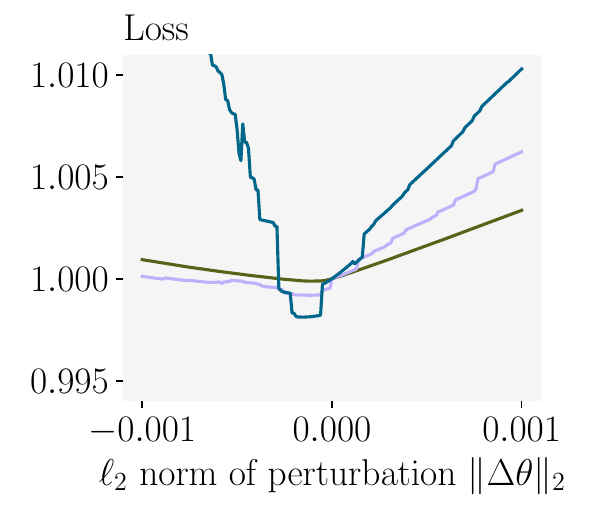}
		\vspace{-6mm}
		\caption{$\sigma=0$ (original)}
		\label{subfig:empirical_smooth1}	
	\end{subfigure}
	\hfil
	\begin{subfigure}[t]{.24\linewidth}
        \includegraphics[width=\linewidth]{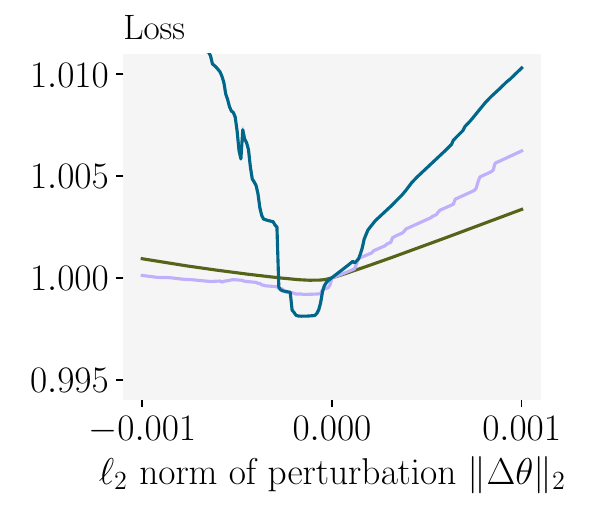}
		\vspace{-6mm}

		\caption{$\sigma=1\times 10^{-6}$}
		\label{subfig:empirical_smooth2}	
	\end{subfigure}
	\hfil
	\begin{subfigure}[t]{.24\linewidth}
        \includegraphics[width=\linewidth]{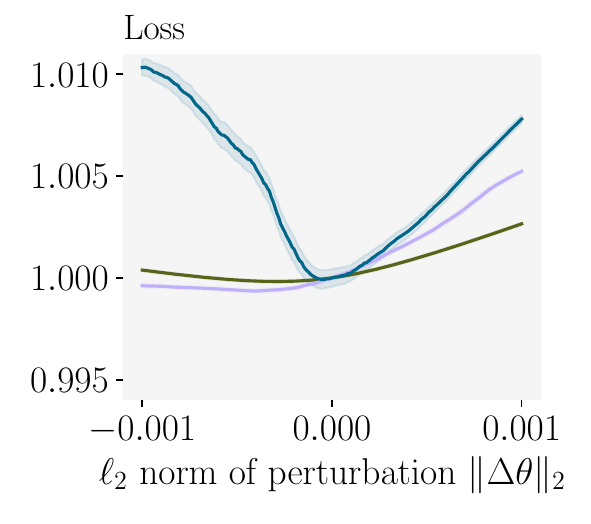}
		\vspace{-6mm}
		\caption{$\sigma=5\times 10^{-5}$}
		\label{subfig:empirical_smooth3}	
	\end{subfigure}
	\hfil
	\begin{subfigure}[t]{.24\linewidth}
        \includegraphics[width=\linewidth]{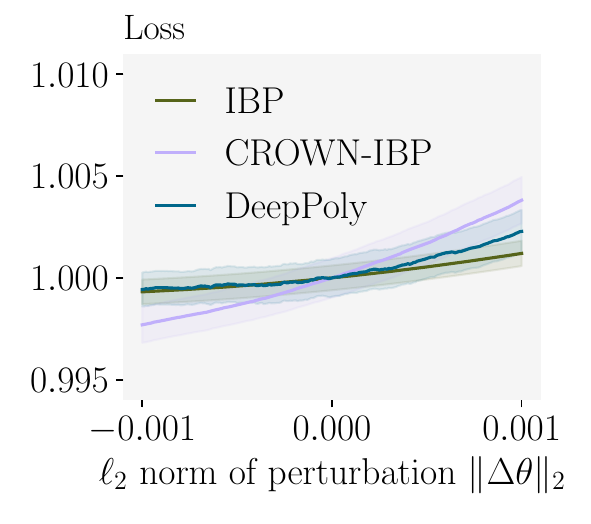}
		\vspace{-6mm}
		\caption{$\sigma=5\times 10^{-4}$}
		\label{subfig:empirical_smooth4}	
	\end{subfigure}
	\vspace{-2mm}
	\caption{The original and Gaussian smoothed loss for different relaxations on a PGD-trained \cnnt, evaluated along the direction of the \deeppoly gradient. Losses are normalized by dividing them by the values at 0, i.e., without perturbation. The smoothed loss is estimated with 128 samples and the corresponding confidence interval is shown as shaded.}
	\label{fig:empirical_smooth}
\end{figure}

\paragraph{Empirical Confirmation}
To empirically confirm that GLS can mitigate discontinuity, non-smoothness, and sensitivity, we plot the original and smoothed loss landscape (along the direction of the \deeppoly gradient) of different relaxations for a \cnnt and different standard deviations in \cref{fig:empirical_smooth}. We normalize all losses by dividing them by their value for the unperturbed weights and estimate the expectation under GLS with sampling.

We observe that the original loss (\cref{subfig:empirical_smooth1}) is discontinuous, non-smooth, and highly sensitive to perturbations for both \crownibp and \deeppoly, consistent with the findings of \citet{JovanovicBBV22} and \citet{LeeLPL21}.
Only the imprecise \ibp loss is continuous and smooth, explaining why the \ibp loss is the basis for many successful certified training methods.
When the loss is smoothed with small standard deviations $\sigma=10^{-6}$ (\cref{subfig:empirical_smooth2}), the local minimum of the \deeppoly loss has a slightly reduced sharpness but is still present. In addition, both the losses for \deeppoly and \crownibp are still highly sensitive. This indicates that $\sigma$ is too small.
When the standard deviation is increased to $\sigma=5\cdot10^{-5}$ (\cref{subfig:empirical_smooth3}), the undesirable local minimum of the \deeppoly loss is removed completely, and both losses become much smoother and less sensitive to perturbations.
However, further increasing the standard deviation to $\sigma=5\cdot10^{-4}$ (\cref{subfig:empirical_smooth4}),
we observe almost flat losses removing the minimum present in the underlying loss, indicating that the smoothing is too strong.

\begin{figure}[t]
	\centering
	\vspace{-2mm}
	\includegraphics[width=.98\linewidth]{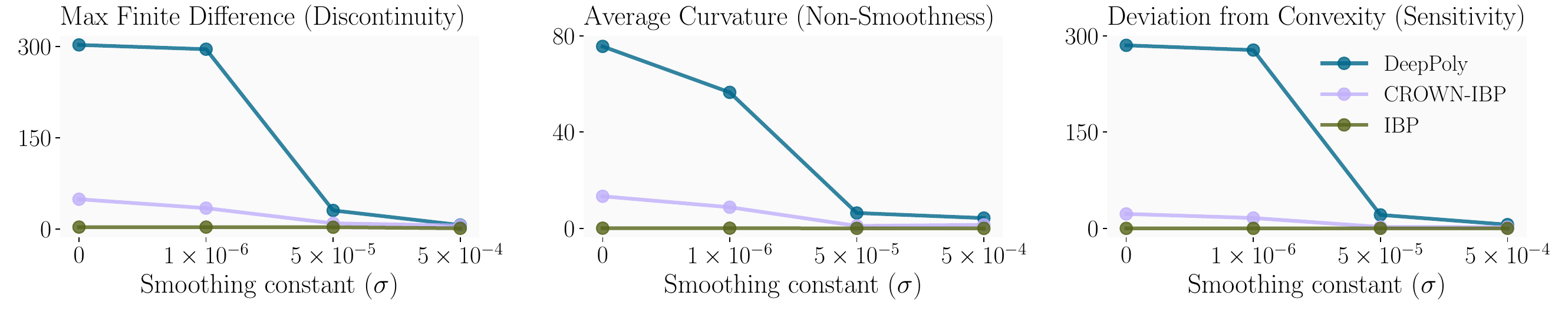}
	\vspace{-2mm}
	\caption{Attenuation of undesirable properties of the loss landscape under increasing smoothing strength. We analyze the loss landscape of the same \cnnt as in \cref{fig:empirical_smooth}. 
	}
	\label{fig:smoothing_scores}
\end{figure}

Moreover, in \cref{fig:smoothing_scores}, we present the evolution of three scores we used as proxies for measuring the three undesirable properties of the loss landscape as we increase the smoothing strength for the same \cnnt network and settings used in \cref{fig:empirical_smooth}. Namely, we compute:
\begin{itemize}[leftmargin=*, itemsep=0pt, topsep=0pt]
	\item the maximum magnitude of finite differences -- defined as $\max_{x} \left|\frac{f(x+h) - f(x)}{h}\right|$ -- as an approximation for the magnitude of the \textit{discontinuities} in the loss function,
	\item the average curvature of the loss function -- defined as the average magnitude of the second order derivative, estimated as $\E_x\left|\frac{f(x-h)-2f(x)+f(x+h)}{h^2}\right|$ -- as a measure of \textit{non-smoothness},
	\item the deviation from convexity -- $D(f)$ as defined in \cref{open_challenges} -- as a proxy for \textit{sensitivity}, 
\end{itemize}
where $f(x)$ represents the normalized loss function w.r.t. the weight perturbation as depicted in \cref{fig:empirical_smooth}, and $h$ is the stepsize used when sampling the loss function for making the plots, roughly equal to 1/50 of the gradient update for a single step of gradient descent. 

We observe that smoothing significantly reduces all of these scores for \deeppoly and CROWN-IBP, thus mitigating the optimization issues induced by the three undesirable properties.

These results empirically confirm the observations in our toy setting and predicted by our theoretical analysis, showing that GLS mitigates the issues related to the paradox of certified training.

Next, in \cref{sec:pgpe} and \cref{sec:rgs}, we show how to apply GLS using PGPE and RGS, respectively. Both PGPE and RGS being estimation-based approximations of GLS, apart from the standard deviation ($\sigma$), they also require the population size ($n_{ps}$) as a hyperparameter to estimate the gradient. 

\subsection{Policy Gradients with Parameter-based Exploration (PGPE)} \label{sec:pgpe}

\begin{figure}[t]
	\centering
	\vspace{-4mm}
	\includegraphics[width=.9\linewidth]{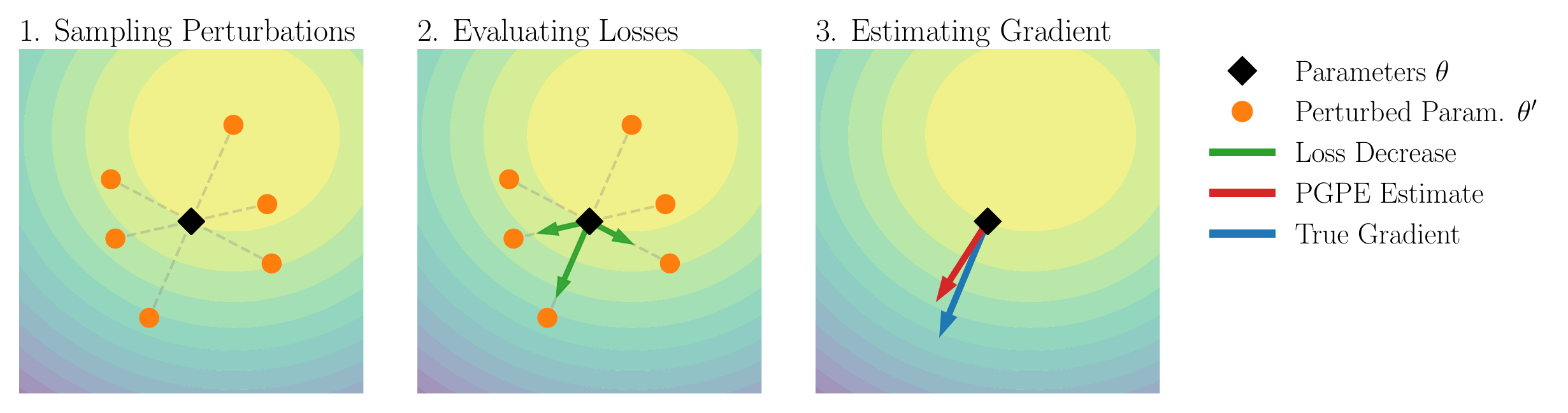}
	\caption{Illustration of \pgpe. (1) We sample random perturbations $\bs{\theta}'$ (\textbf{\textcolor{orange}{orange}} circles) symmetrically around the current parameters $\bs{\theta}$ (\textbf{black} diamond) from a Gaussian distribution $\gN({\bs 0}, {\bs \sigma})$. (2) For each pair of perturbations, we evaluate the loss and compute directional differences (\textbf{\textcolor{my-full-green}{green}} arrows, longer arrows represent larger differences). (3) The PGPE gradient estimate (\textbf{\textcolor{my-full-red}{red}} arrow) is computed as a sum of all sampled directions weighted by the respective observed loss differences. The result is an approximation of the true gradient of the smoothed loss function (\textbf{\textcolor{my-full-blue}{blue}} arrow).
	 }
	\vspace{-2mm}
	\label{fig:pgpe_demo}
\end{figure}

PGPE \citep{SehnkeORGPS10} is a gradient-free optimization algorithm that optimizes the Gaussian Smoothed loss $L_\sigma(\bs{\theta}) \coloneqq \E_{\bs{\theta}' \sim \mathcal{N}(\bs{\theta}, \sigma^2 \bs{I}) } L(\bs{\theta'})$ using a zeroth-order method, where the loss is not evaluated at a single parameterization of the network, but rather at a (normal) distribution of parameterizations.

\pgpe samples $n=n_{ps}/2$ weight perturbations $\bs{\epsilon}_i \sim \mathcal{N}({\bs 0}, \bs{\sigma}^2)$, and evaluates the loss on $\bs{\theta} + \bs{\epsilon_i}$, and $\bs{\theta} - \bs{\epsilon_i}$, computing $r^{+}_{i} = L(\bs{\theta} + \bs{\epsilon_i})$ and $r^{-}_{i} = L(\bs{\theta} - \bs{\epsilon}_i)$.
These pairs of symmetric points are then used to compute gradient estimates with respect to both the mean of the weight distribution $\bs{\theta}$ and its standard deviation $\bs{\sigma}$:
$\nabla_{\bs{\theta}} \hat{L}_{\bs{\sigma}}(\bs{\theta}) \propto \sum_i \bs{\epsilon}_i (r^{+}_i - r^{-}_i)$
and $\nabla_{\bs{\sigma}} \hat{L}_{\bs{\sigma}}(\bs{\theta}) \propto \sum_i \left(\tfrac{r^+_i + r^-_i}{2} -b\right)\tfrac{\bs{\epsilon}_i^2 - \bs{\sigma}^2}{\bs{\sigma}},$
where $b = \frac{1}{2 n} \sum_i \left( r^+_i + r^-_i \right)$ is called baseline loss and is the average of loss values over all $2n$ samples. \cref{fig:pgpe_demo} visualizes such a gradient estimate. The gradient approximations $\nabla_{\bs{\theta}} \hat{L}_{\bs{\sigma}}(\bs{\theta})$ and $\nabla_{\bs{\sigma}} \hat{L}_{\bs{\sigma}}(\bs{\theta})$ are used to update the mean weights $\bs{\theta}$ and the standard deviation $\bs{\sigma}$, respectively. 
By design, \pgpe approximately optimizes the Gaussian smoothed loss \citep{SehnkeORGPS10}.

As \emph{no} backward propagation is needed to compute these gradient estimates, \pgpe is comparable to neuro-evolution algorithms. In this context, it is among the best-performing methods for supervised learning \citep{lange2023neuroevobench}. This property also allows us to apply it for training with tighter, but non-differentiable bounding methods, such as $\alpha$-CROWN \citep{xu2020fast}.

\begin{figure}[t]
	\centering
	\includegraphics[width=.9\linewidth]{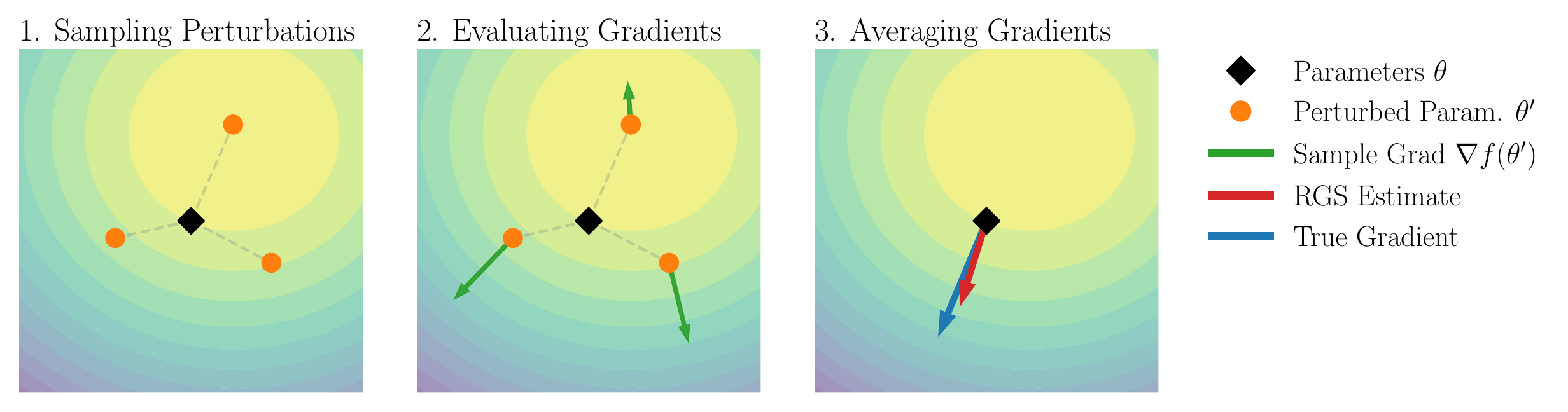}
	\caption{Illustration of \rgs. (1) We sample random perturbations $\bs{\theta}'$ (\textbf{\textcolor{orange}{orange}} circles) around the current parameters $\bs{\theta}$ (\textbf{black} diamond) from a Gaussian distribution $\gN({\bs 0}, {\bs \sigma})$. (2) For each perturbation, we evaluate the gradient (\textbf{\textcolor{my-full-green}{green}} arrows). (3) The RGS gradient estimate (\textbf{\textcolor{my-full-red}{red}} arrow) is computed as the average of the sampled gradients. The result approximates the true gradient of the smoothed loss function (\textbf{\textcolor{my-full-blue}{blue}} arrow).
	}
	\vspace{-4mm}
	\label{fig:rgs_demo}
\end{figure}

\subsection{Randomized Gradient Smoothing (RGS)}
\label{sec:rgs}

While the loss smoothing induced by the sampling procedure of \pgpe leads to a provably continuous and infinitely differentiable loss surface, it can be costly to compute. To reduce the training costs, we propose to approximate GLS by RGS \citep{duchi2012randomized}. RGS approximates the gradient of the smoothed loss by sampling $n_{ps}$ perturbations $\bs{\epsilon}_i \sim \mathcal{N}({\bs 0}, \bs{\sigma}^2)$ and then averaging the gradients of the loss function for the perturbed network weights $\bs{\theta} + \bs{\epsilon}_i$:
\begin{align}
	\nabla_\theta \hat{L}_\sigma(\theta) \propto \frac{1}{n_{ps}}\sum_i \nabla_\theta {L}(\theta+\epsilon_i).
\end{align}
While RGS, which approximates in first-order does not provably recover in expectation the gradient of the smoothed loss when the original function is discontinuous (see \cref{sec:rgs-properties}), \citet{duchi2012randomized} have shown its empirical effectiveness, even with a tiny sample size ($n_{ps}=2$). Therefore, we apply this alternative to study the performance of GLS in larger networks, as RGS requires much fewer samples than PGPE and thus scales better.
Further, contrary to before, $\sigma$ is now a hyperparameter that needs to be tuned rather than learned.
A comparison of training costs is included in \cref{app:train_costs}, where RGS is shown to be up to 40 times faster than PGPE.

\section{Experimental Evaluation}

We now extensively evaluate the effect of GLS via \pgpe and RGS on the training characteristics of different relaxation methods. First, we show in \cref{sec:main_results} that \pgpe enables training with tight relaxations, even when the relaxation is not differentiable. Second, we demonstrate in \cref{sec:deeper_nets} that \rgs scales GLS training to deeper networks, surpassing the performance of the SOTA methods on the same network architecture in many settings. The impact of different hyperparameters on the performance of the proposed methods is studied in \cref{sec:ablations}, and a comparison of \pgpe and \rgs is provided in \cref{app:pgpe-rgs-comparison}. Overall, our results show that GLS can enable certified training with tight relaxations.

\subsection{Experimental Setup}
We implement all certified training methods in PyTorch \citep{PaszkeGMLBCKLGA19} and conduct experiments on \mnist \citep{lecun2010mnist}, \cifar \citep{krizhevsky2009learning} and \TIN \citep{Le2015TinyIV} using $l_\infty$ perturbations and versions of the \cnnt and \cnnf architectures (see \cref{tab:architectures} in \cref{app:train_details}). For more details on the experimental setting including all hyperparameters, see \cref{app:train_details}.

\paragraph{Standard Certified Training}
For standard certified training using back-propagation (referred to below as \grad for clarity), we use similar hyperparameters as in the literature and initialize all models using the IBP initialization proposed by \citet{ShiWZYH21}. In particular, we also use the Adam optimizer \citep{KingmaB14}, follow their learning rate and $\epsilon$-annealing schedule, use the same batch size and gradient clipping threshold, and use the same $\eps$ for training and certification in all settings. For the state-of-the-art methods \sabr \citep{MullerE0V23}, \staps \citep{mao2023connecting}, and MTL-\ibp \citep{palma2024expressive}, we conduct an extensive optimization of their network-specific hyperparameters and only report the best results.

\paragraph{\pgpe Training}
We train our \pgpe models using the multi-GPU, multi-actor implementation from \texttt{evotorch} \citep{toklu2023evotorch}.
As \pgpe training is computationally expensive, we initialize from an adversarially trained (PGD, \citep{MadryMSTV18}) model. This can be seen as a warm-up stage as is common also for other certified training methods \citep{ShiWZYH21,MullerE0V23,mao2023connecting}. We only use $\epsilon$-annealing for the larger perturbation magnitudes on both \mnist and \cifar and choose the learning rate based on stability at the beginning of the training. Unless indicated otherwise, we run the PGPE algorithm with a population size of $n_{ps} = 256$ and an initial standard deviation for weight sampling of $\sigma_\pgpe = 10^{-3}$.

\paragraph{RGS Training}
We train our RGS models using the same hyperparameters as for \grad training. We use a population size of $n_{ps} = 2$ and an initial standard deviation of $\sigma_\rgs = 10^{-3}$. As RGS does not dynamically adjust the standard deviation, we choose to decay it at the same time steps as the learning rate. More details about the hyperparameters used can be found in \cref{app:train_details}.

\paragraph{Certification}
We use the state-of-the-art complete verification method \mnbab \citep{FerrariMJV22} with the same settings as used by \citet{MullerE0V23} for all networks independently of the training method. We note that this is in contrast to \citet{JovanovicBBV22} who used the same relaxation for training and verification. By doing this, we aim to assess true robustness regardless of the tightness of different relaxations.

\subsection{GLS Enables Training with Tight Relaxations} \label{sec:main_results}
We first compare the performance of training with various differentiable convex relaxations using either standard backpropagation (\grad) or GLS-based methods (\pgpe and \rgs). The result is shown in \cref{tab:main_results}.

\begin{table*}[t]
	\centering	
		\vspace{-5mm}
		\caption{Comparison of the Natural, Certified and Adversarial Accuracies of \cnnt networks trained using different convex relaxations on the \mnist and \cifar datasets for different perturbation sizes ($\epsilon_\infty$). We compare GLS-based algorithms PGPE \citep{SehnkeORGPS10} and RGS \citep{duchi2012randomized} with the baseline approach (gradients obtained directly by backpropagation -- GRAD). We use the state-of-the-art method \mnbab \citep{FerrariMJV22} for certification.}
		\resizebox{0.98\linewidth}{!}{
		\begin{threeparttable}
			\footnotesize
			\begin{tabular}{@{}lclccccccccc@{}}
				\toprule
				\multirow{2.5}{*}{Dataset} & \multirow{2.5}{*}{$\epsilon_\infty$} & \multirow{2.5}{*}{\makecell[l]{Convex \\ Relaxation}} & \multicolumn{3}{c}{Natural Accuracy [\%]} & \multicolumn{3}{c}{Certified Accuracy [\%]}  & \multicolumn{3}{c}{Adversarial Accuracy [\%]} \\
                \cmidrule(rl){4-6} \cmidrule(rl){7-9} \cmidrule(l){10-12}
                &&& \textsc{GRAD} & \textsc{PGPE} & \textsc{RGS} & \textsc{GRAD} & \textsc{PGPE} & \textsc{RGS} & \textsc{GRAD} & \textsc{PGPE} & \textsc{RGS} \\
				\midrule
				\multirow{8.5}{*}{MNIST}    & \multirow{4}{*}{0.1} 
				  & \ibp               & 96.02 & 94.52   & 96.13 & \textbf{91.23}  & 87.02 	        & 91.77          & 91.23 & 87.03 & 91.77  \\
				& & \hboxp             & 94.79 & 96.12   & 95.52 & 88.18 		   & 90.57    	    & 90.13          & 88.18 & 90.58 & 90.15  \\
				& & \crownibp          & 94.33 & 96.69   & 96.74 & 88.76 		   & 90.23   	    & 91.05          & 88.77 & 90.25 & 91.10  \\
				& & \deeppoly          & 95.95 & 97.44   & 97.37 & 90.04 		   & \textbf{91.53} & \textbf{91.88} & 90.08 & 91.79 & 92.03  \\
				\cmidrule(l){2-12}
				& \multirow{4}{*}{0.3}     
				  & \ibp               & 91.02 & 89.16   & 91.99 & \textbf{77.23}  & 74.00    		& \textbf{77.07} & 77.27 & 74.08 & 77.15  \\
				& & \hboxp             & 83.75 & 86.58   & 83.81 & 57.86		   & 70.52    		& 58.37          & 57.92 & 70.66 & 58.69  \\
				& & \crownibp          & 86.97 & 90.57   & 88.86 & 70.55		   & 71.95    		& 71.91          & 70.56 & 72.24 & 71.94  \\
				& & \deeppoly          & 85.70 & 91.05   & 88.51 & 66.69		   & \textbf{74.28} & 71.36          & 66.70 & 74.98 & 71.49  \\
				\midrule
				\multirow{6.5}{*}{CIFAR-10} & \multirow{3}{*}{2/255}   
				  & \ibp               & 48.05 & 44.55   & 47.70 & \textbf{37.69}  & 34.09    		& 37.28          & 37.70 & 34.10 & 37.28  \\
				& & \crownibp          & 44.49 & 51.19   & 53.74 & 35.75 		   & 37.51    		& 41.00          & 35.75 & 37.65 & 41.46  \\
				& & \deeppoly          & 47.70 & 54.17   & 54.93 & 36.72		   & \textbf{38.95} & \textbf{41.14} & 36.72 & 40.20 & 42.03  \\
				\cmidrule(l){2-12}
				& \multirow{3}{*}{8/255}   
				  & \ibp               & 34.63 & 30.48   & 33.23 & \textbf{25.72}  & 21.75    		& \textbf{24.56} & 25.74 & 21.75 & 24.58  \\
				& & \crownibp          & 31.60 & 32.36   & 35.17 & 22.66		   & 21.40    		& 23.92          & 22.66 & 21.42 & 24.18  \\
				& & \deeppoly          & 33.06 & 31.37   & 35.61 & 22.97		   & \textbf{22.19} & 23.81          & 22.98 & 22.19 & 24.21  \\
				\bottomrule
			\end{tabular}
			\label{tab:main_results}
		\end{threeparttable}
		}
\end{table*}

\paragraph{\grad Training}
We train the same \cnnt on \mnist and \cifar at the established perturbation magnitudes using standard certified training with \ibp, \hboxp, \crownibp, and \deeppoly.
We observe that across all these settings \ibp dominates the other methods both in terms of standard and certified accuracy, confirming the paradox of certified training.
Specifically, \hboxp, \crownibp, and \deeppoly tend to perform similarly, with \crownibp being significantly better at \mnist $\eps = 0.3$, indicating that when the loss is discontinuous, non-smooth and sensitive, tightness of the training relaxation is less relevant.

\paragraph{\pgpe Training}
Training the same \cnnt with \pgpe in the same settings we observe that the performance ranking changes significantly (see \cref{tab:main_results}).
Now, training with \ibp performs strictly worse than training with \deeppoly across all datasets and perturbation sizes.
In fact, the more precise \deeppoly bounds now yield the best certified accuracy across all settings, even outperforming \grad-based training methods at low perturbation radii.
Interestingly, \ibp still yields better certified accuracy at large perturbation radii than \hboxp and \crownibp, although at significantly worse natural accuracies. This is likely because more severe regularization is required in these settings. For a more detailed discussion on the issue of certified training for large perturbations see \cref{app:discussion}.

While \deeppoly + \pgpe outperforms \deeppoly + \grad in almost all settings in \cref{tab:main_results} on the same network architecture, sometimes by a wide margin, it does not reach the general SOTA results of classic and heavily optimized \grad training methods. We believe this is caused by three key factors: 
First, \pgpe computes a gradient approximation in an $\tfrac{n_\text{ps}}{2}$-dimensional subspace. 
To cover the full parameter space, we would need the population size $n_\text{ps}$ to be twice the number of network parameters, which is computationally intractable even for small networks.
Thus, we only get low-dimensional gradient approximations, slowing down training (see \cref{tab:popsize-ablation} and \cref{fig:train_dynamics}). 
Second, again due to the high cost of training with \pgpe, we used relatively short training schedules and were unable to optimize hyperparameters for the different settings.
Finally, \pgpe-based certified training is less optimized, compared to standard certified training which has been extensively optimized over the past years \citep{ShiWZYH21,MullerE0V23,palma2024expressive}.

\paragraph{\rgs training}
When applying \rgs training to the same \cnnt architecture, we observe that \rgs significantly improves the performace of training with tighter relaxations in all settings. In particular, \deeppoly + \rgs outperforms all other methods in the case of small perturbations, while \ibp-\grad is still the best method for large perturbations. 
We note that this is potentially because RGS, as a first-order approximation of GLS, does not necessarily enjoy the continuity that GLS brings.
Still, the performance improvements point toward the potential of \rgs to alleviate the issues of tight relaxations, while also being able to scale to deeper networks, as we show in \cref{sec:deeper_nets}.

\vspace{-1mm}
\paragraph{Takeaway:}
GLS-based methods like PGPE and RGS enable effective training with tight convex relaxations, overcoming the paradox where looser bounds (e.g., IBP) typically outperform tighter ones. PGPE boosts certified accuracy with DeepPoly but is limited by its costly and imperfect gradient estimation, while RGS offers scalable improvements, especially for small perturbations, highlighting GLS's potential for stable and tight certified training.

\begin{wraptable}{r}{.45\textwidth}
    \centering
    \vspace{-1.4em}
    \caption{Accuracies of \cnnty on MNIST $\epsilon=0.1$ trained with different algorithms.}
    \label{tab:nondiff-results}
    \centering
    \footnotesize
    \resizebox{0.95\linewidth}{!}{
        \begin{tabular}{lccc}
        \toprule
        Method        & Nat. [\%] & Cert. [\%] & Adv. [\%] \\
        \midrule
        \ibp-GRAD             & 89.76 & 82.46 & 82.48 \\
        \deeppoly-GRAD        & 91.27 & 82.04 & 82.05 \\
        \deeppoly-PGPE        & 91.94 & 85.00 & 85.04 \\
        $\alpha$-CROWN-PGPE & \textbf{92.15} & \textbf{85.15} & \textbf{85.17} \\
        \bottomrule
        \end{tabular}
    }
    \vspace{-3mm}
\end{wraptable}

\vspace{-1mm}
\subsection{\pgpe enables non-differentiable relaxations}

Next, we show that \pgpe has a unique benefit in that it allows training with non-differentiable relaxations, which we demonstrate by training with the non-differentiable $\alpha$-CROWN relaxation. 
Since $\alpha$-CROWN is even more expensive than \deeppoly, we train it with a smaller version of \cnnt called \cnnty and set the number of iterations in $\alpha$-\crown slope optimization to be merely 1. \cref{tab:nondiff-results} shows that training with $\alpha$-CROWN-\pgpe further improves the certified accuracy compared to training with \deeppoly-\pgpe. This confirms that \pgpe can be used to train with non-differentiable relaxations, resulting in even better robustness-accuracy trade-offs. We remark that \pgpe is not limited to $\alpha$-CROWN, but can be used with any non-differentiable relaxation, including those relying on branch and bound-based procedures or multi-neuron constraints. Although these methods are computationally expensive and thus may be only applied in training small networks, they are particularly useful in safety-critical applications such as aircraft control \citep{owen2019acas} or embedded medical devices \citep{shoeb2009micropower}, where models are usually even smaller.

\begin{table}[ht]
    \centering
    \caption{Comparison between networks trained with \deeppoly-RGS, \crownibp-RGS and SOTA GRAD methods on small perturbation settings. The best performance for each dataset and architecture is \textbf{highlighted}. Numbers in \textit{italic} represent results for GRAD methods obtained on the SOTA \cnns architecture, which is more than 10 times larger than the \cnnf and \cnnfl architectures.}
    \label{tab:cnn5-results}
    \centering

    \footnotesize
    \resizebox{0.8\linewidth}{!}{
		\begin{tabular}{@{}cclccc@{}}
            \toprule
            {Dataset} & \makecell{Network \\ (params.)}  & Method  & \makecell{Nat. Acc. \\ $\text{[\%]}$}    & \makecell{Cert. Acc. \\ $\text{[\%]}$} & \makecell{Adv. Acc. \\ $\text{[\%]}$}  \\
            \midrule
            \multirow{13}{*}{\makecell{\mnist \\ $\epsilon_\infty=0.1$}}     
            & \multirow{8}{*}{\makecell{\cnnf \\ (166K)}}   
              & IBP     & 97.94          & 95.82          & 95.83          \\
            & & SABR    & 98.81          & 96.28          & 96.31          \\
            & & STAPS   & 98.74          & 96.05          & 96.09          \\
            & & MTL-IBP & 98.74          & 96.25          & 96.29          \\
            & & CROWN-IBP      & 98.19          & 95.42          & 95.42          \\
            & & CROWN-IBP-RGS  & 98.43          & 95.64          & 95.65          \\
            & & \deeppoly      & 98.50          & 95.95          & 95.97          \\
            & & \deeppoly-RGS  & \textbf{98.97} & \textbf{97.15} & \textbf{97.16} \\
            \cmidrule(rl){2-6}
            & \multirow{2}{*}{\makecell{\cnnfl \\ (1.25M)}}  
              & MTL-IBP        & 98.91          & 97.17          & 97.33          \\
            & & \deeppoly-RGS  & \textbf{99.21} & \textbf{97.61} & \textbf{97.76} \\
            \cmidrule(rl){2-6}
            & \multirow{2}{*}{\makecell{\cnns \\ (13.3M)}}   
              & IBP           & \textit{98.87} & \textit{98.26} & \textit{98.27} \\
            & & TAPS          & \textit{99.16} & \textit{98.52} & \textit{98.58} \\
            \midrule
            \multirow{12}{*}{\makecell{\cifar \\ $\epsilon_\infty=2/255$}}      
            & \multirow{7}{*}{\makecell{\cnnf \\ (281K)}}   
              & IBP     & 54.92          & 45.36          & 45.36          \\
            & & SABR    & 66.73          & 52.11          & 52.55          \\
            & & MTL-IBP & 67.03          & 53.81          & 55.18          \\
            & & CROWN-IBP      & 60.91          & 49.45          & 49.68          \\
            & & CROWN-IBP-RGS  & 63.22          & 50.73          & 51.18          \\
            & & \deeppoly      & 65.43          & 53.16          & 54.10          \\
            & & \deeppoly-RGS  & \textbf{67.88} & \textbf{54.91} & \textbf{56.12} \\
            \cmidrule(rl){2-6}
            & \multirow{2}{*}{\makecell{\cnnfl \\ (1.25M)}}  
            & MTL-IBP        & 70.60          & 56.36          & 59.05          \\
            & & \deeppoly-RGS  & \textbf{72.64} & \textbf{59.34} & \textbf{61.23} \\
            \cmidrule(rl){2-6}
            & \multirow{2}{*}{\makecell{\cnns \\ (17.2M)}}   
            & IBP           & \textit{67.49} & \textit{55.99} & \textit{56.10} \\
            & & MTL-IBP       & \textit{78.82} & \textit{64.41} & \textit{67.69} \\
            \midrule
            \multirow{7.5}{*}{\makecell{\TIN \\ $\epsilon_\infty=1/255$}}        
            & \multirow{5}{*}{\makecell{\cnnf \\ (1.17M)}}   
              & IBP            & 19.55          & 13.92          & 13.93          \\
            & & MTL-IBP        & 26.92          & 18.07          & 18.16          \\
            & & CROWN-IBP-LF      & 21.91          & 16.43          & 16.43          \\
            & & CROWN-IBP-LF-RGS  & 22.97          & 16.89          & 16.89          \\
            & & \deeppoly-RGS  & \textbf{27.84} & \textbf{19.73} & \textbf{20.40} \\
            \cmidrule(rl){2-6}
            & \multirow{2}{*}{\makecell{\cnns \\ (17.3M)}}   
              & IBP     & \textit{26.77} & \textit{19.82} & \textit{19.84} \\
            & & MTL-IBP & \textit{35.97} & \textit{27.73} & \textit{28.49} \\
            \bottomrule
        \end{tabular}
    }
\end{table}

\vspace{-1mm}
\subsection{RGS Scales GLS Training}
\label{sec:deeper_nets}
\vspace{-1mm}

We have demonstrated the empirical advantages of GLS instantiated with PGPE. 
However, as PGPE is computationally expensive and limited to small models, more scalable methods are required to train larger networks. In this section, we extensively evaluate RGS, showing that its efficiency allows us to scale to larger models, surpassing the performance of the SOTA methods on the same network architecture on standard evaluation settings when $\epsilon_\infty$ is relatively small.

RGS overcomes the low-rank gradient and computational cost issues of PGPE: even with a small population size (hence low training costs), we obtain full-rank gradient approximations, enabling faster and better optimization and allowing us to even scale our experiments to \TIN. We analyze the results of training with RGS on the \cnnf and \cnnfl (a wider version of \cnnf) architectures and compare them with \ibp and the SOTA \grad-based methods  \citep{mao2024ctbench} trained on \cnns in \cref{tab:cnn5-results}.
Encouragingly, RGS significantly boosts the performance of \deeppoly training. We observe that \deeppoly + RGS dominates all other methods, substantially improving even over state-of-the-art \grad-based methods with hyperparameters fine-tuned on \cnnf and \cnnfl. Further, the performance of \deeppoly + RGS on the small \cnnf becomes comparable to the performance of \grad-\ibp on the much larger \cnns architecture used by recent SOTA methods, and the \cnnfl trained with \deeppoly-RGS exceeds the performance of \cnns trained with IBP by a large margin. These results agree well with our expectation that bound tightness becomes increasingly important with network depth, as overapproximation errors can grow exponentially with depth \citep{ShiWZYH21,MullerE0V23,MaoMFV24}.  We remark that scaling \deeppoly-\rgs to \cnns used by the SOTA methods is still infeasible due to the high computational cost of evaluating \deeppoly (RGS only doubles the cost!), but we show that RGS can still be used with the cheaper \crownibp relaxation on this architecture in \cref{tab:cnn7-results} in \cref{app:crownibp-cnn7}.

\subsection{RGS Training in Large Perturbation Settings} 
In \cref{tab:cnn5-results-large} we provide experimental data for training \cnnf networks using \deeppoly + RGS. We observe that while \deeppoly + RGS manages to obtain similar natural accuracies with gradient-based \ibp, the certified accuracies are significantly lower. This is likely because to gain certifiability for the large epsilon settings the networks require a stronger regularisation than the \deeppoly relaxation can provide.

\begin{table}[ht]
    \centering
    \vspace{-1mm}
    \caption{Accuracies of a \cnnf depending on training method.}
    \label{tab:cnn5-results-large}
    \centering
    \footnotesize
    \resizebox{0.7\linewidth}{!}{
		\begin{tabular}{clccc}
            \toprule
            Dataset                    & Method       & Nat. [\%]      & Cert. [\%]     & Adv. [\%]       \\
            \midrule
            \multirow{5}{*}{\makecell{\mnist \\ $\epsilon_\infty=0.3$}} 
            & IBP (used as init) & 94.95          & 87.71          & 87.80          \\
            & SABR         & \textbf{97.78} & 88.26          & \textbf{89.33} \\
            & MTL-IBP      & 97.08          & 88.68          & 88.95          \\
            \cmidrule(rl){2-2}
            & \deeppoly-RGS       & 95.79          & 87.04          & 87.17          \\
            & \deeppoly-RGS (IBP) & 95.47          & \textbf{88.69} & 88.79          \\
            \midrule
            \multirow{5}{*}{\makecell{\cifar \\ $\epsilon_\infty\!=\!8/255$}}     
            & IBP (used as init) & 41.05          & 29.12          & 29.14          \\
            & SABR         & 43.30          & 29.50          & 29.55          \\
            & MTL-IBP      & \textbf{44.53} & \textbf{29.62} & \textbf{29.73} \\
            \cmidrule(rl){2-2}
            & \deeppoly-RGS       & 40.10          & 25.25          & 25.93          \\
            & \deeppoly-RGS (IBP) & 41.66          & 29.25          & 29.31          \\
            \bottomrule
            \end{tabular}
    }
    \vspace{-1mm}
\end{table}

This is in agreement with the findings of \citet{mao2024ctbench}, which after extensive hyperparameter tuning, show that IBP trained networks can obtain very close performance to SOTA methods in the large perturbation settings. For example, while the SOTA method, MTL-IBP, improves IBP by more than $10\%$ for CIFAR-10, $\epsilon=2/255$, it merely improves IBP by $0.13\%$ for CIFAR-10, $\epsilon=8/255$, representing a roughly 100x reduction in relative improvement. We observe a similar pattern in our experiments with standard certified training methods on \cnnf. A more detailed discussion can be found in \cref{app:discussion}.

To verify that training with tighter relaxations can still yield improvements in the large perturbation settings, we initialize the \cnnf networks with \ibp-trained weights and further train them with \deeppoly + RGS. The results are shown in \cref{tab:cnn5-results-large}, denoted by \deeppoly-RGS (IBP). We observe that training with \deeppoly + RGS increases both natural and certified accuracies when compared to the \ibp-trained initialization, with certified accuracy reaching a similar level with MTL-IBP on \mnist 0.3. However, on \cifar $\epsilon=8/255$, this is still weaker than the SOTA MTL-IBP, although it strictly improves over IBP. This demonstrates the ability of tighter relaxations to still improve training in the large perturbation settings, but more work is needed to surpass the performance of SOTA methods.

\section{Discussion and Limitations}
\label{sec:limitations}

This work shows the promise of Gaussian Loss Smoothing (GLS) to enable certified training with tight relaxations. PGPE and RGS, our proposed methods implementing GLS, achieve strong performance empirically. However, there are several limitations and challenges that need to be addressed in future work. First, GLS provably mitigates the discontinuity, non-smoothness, and perturbation sensitivity issues identified, but it is unknown whether these are all the factors contributing to the paradox of certified training. Future work should investigate other potential factors and how they can be addressed. Second, while our methods achieve strong performance, they are computationally expensive. Future work should focus on more computationally efficient smoothing approaches. Finally, we present a first step towards training with tight relaxations, but our methods could be further optimized, similar to how \ibp-based methods have been optimized over the years. Overall, our work opens up a new direction for certified training using tight relaxations, and we hope it will inspire future work in this area.

\section{Related Work}
\label{sec:related_work_real}

To guarantee robustness of neural networks, two main approaches have been proposed. Randomized smoothing \citep{CohenRK19,SalmanLRZZBY19,SalmanSYKK20,JeongPKLKS21,HorvathM0V22,VaishnaviER22,JeongKS23,abs-2410-06895} applies random perturbation on the input and draw statistical robustness guarantees on the smoothed classifier. To provide deterministic guarantees, convex relaxations of the neural network are commonly adopted \citep{MirmanGV18,WongSMK18,SinghGMPV18,ZhangWCHD18,SinghGPV19}. These methods are limited in completeness \citep{MirmanBV22,baader2024expressivity,mao2025multineuron}, thus the most effective certification algorithms combine branch-and-bound with convex relaxations to yield complete certification \citep{BunelHBD20,XuZ0WJLH21,WangZXLJHK21,FerrariMJV22}. However, these methods are computationally expensive and do not scale well to large networks.
Thus, neural networks tailored for certification have been widely studied \citep{GowalIBP2018,BalunovicV20,ZhangCXGSLBH20,ShiWZYH21,PalmaIBPR22,MullerE0V23,mao2023connecting,MaoMFV24,palma2024expressive}. Surprising phenomena, however, have been observed when training certified models, where the performance of certified models trained with more accurate bounds may not exceed that of models trained with less accurate bounds. While some reasons have been identified \citep{JovanovicBBV22,LeeLPL21}, it is not yet clear how to solve this issue. This work examines whether solving the identified optimization difficulties can help solve this problem.

\section{Conclusion}\label{sec:conclusion}
This work shows that the three issues contributing to the paradox of certified training identified by prior works, namely discontinuity, non-smoothness, and perturbation sensitivity, can be mitigated by Gaussian Loss Smoothing (GLS), based on sound theoretical analyses. We instantiate GLS with two methods: Policy Gradients with Parameter-based Exploration (PGPE) and Randomized Gradient Smoothing (RGS). Empirically, we demonstrate that both improve training with tight relaxations, presenting a solid step towards overcoming the paradox. Further, we show that both methods have unique advantages: PGPE allows training with non-differentiable relaxations, while RGS scales better. Our results confirm the importance of loss continuity, smoothness, and insensitivity in certified training, and pave the way for future work to leverage tighter relaxations for certified training.

\section*{Reproducibility Statement}

We release the complete code used for our experiments at \href{https://github.com/stefanrzv2000/GLS-Cert-Training}{github.com/stefanrzv2000/GLS-Cert-Training}. 
A detailed description of the experimental setup and hyperparameters is provided in \cref{app:train_details}.

\section*{Acknowledgements}

This research was partially funded by the Ministry of Education and Science of Bulgaria (support for INSAIT, part of the Bulgarian National Roadmap for Research Infrastructure).

This work has been done as part of the EU grant ELSA (European Lighthouse on Secure and Safe AI, grant agreement no. 101070617) and the SERI grant SAFEAI (Certified Safe, Fair and Robust Artificial Intelligence, contract no. MB22.00088). Views and opinions expressed are however those of the authors only and do not necessarily reflect those of the European Union or European Commission. Neither the European Union nor the European Commission can be held responsible for them. 

The work has received funding from the Swiss State Secretariat for Education, Research and Innovation (SERI).

\message{^^JLASTBODYPAGE \thepage^^J}

\bibliography{references}

\begin{thebibliography}{60}
\providecommand{\natexlab}[1]{#1}
\providecommand{\url}[1]{\texttt{#1}}
\expandafter\ifx\csname urlstyle\endcsname\relax
  \providecommand{\doi}[1]{doi: #1}\else
  \providecommand{\doi}{doi: \begingroup \urlstyle{rm}\Url}\fi

\bibitem[Baader et~al.(2024)Baader, Mueller, Mao, and
  Vechev]{baader2024expressivity}
Maximilian Baader, Mark~Niklas Mueller, Yuhao Mao, and Martin Vechev.
\newblock Expressivity of re{LU}-networks under convex relaxations.
\newblock In \emph{Proc. ICLR}, 2024.

\bibitem[Balunović \& Vechev(2020)Balunović and Vechev]{BalunovicV20}
Mislav Balunović and Martin~T. Vechev.
\newblock Adversarial training and provable defenses: Bridging the gap.
\newblock In \emph{Proc. of ICLR}, 2020.

\bibitem[Biggio et~al.(2013)Biggio, Corona, Maiorca, Nelson, Srndic, Laskov,
  Giacinto, and Roli]{BiggioCMNSLGR13}
Battista Biggio, Igino Corona, Davide Maiorca, Blaine Nelson, Nedim Srndic,
  Pavel Laskov, Giorgio Giacinto, and Fabio Roli.
\newblock Evasion attacks against machine learning at test time.
\newblock In \emph{Proc of ECML PKDD}, 2013.
\newblock \doi{10.1007/978-3-642-40994-3_25}.

\bibitem[Brix et~al.(2023)Brix, M{\"{u}}ller, Bak, Johnson, and
  Liu]{BrixMBJL23}
Christopher Brix, Mark~Niklas M{\"{u}}ller, Stanley Bak, Taylor~T. Johnson, and
  Changliu Liu.
\newblock First three years of the international verification of neural
  networks competition {(VNN-COMP)}.
\newblock \emph{CoRR}, abs/2301.05815, 2023.
\newblock \doi{10.48550/ARXIV.2301.05815}.

\bibitem[Bunel et~al.(2020)Bunel, Hinder, Bhojanapalli, and
  Dvijotham]{BunelHBD20}
Rudy Bunel, Oliver Hinder, Srinadh Bhojanapalli, and Krishnamurthy Dvijotham.
\newblock An efficient nonconvex reformulation of stagewise convex optimization
  problems.
\newblock In \emph{Proc. of NeurIPS}, 2020.

\bibitem[Cohen et~al.(2019)Cohen, Rosenfeld, and Kolter]{CohenRK19}
Jeremy~M. Cohen, Elan Rosenfeld, and J.~Zico Kolter.
\newblock Certified adversarial robustness via randomized smoothing.
\newblock In \emph{Proc. of ICML}, 2019.

\bibitem[Croce \& Hein(2020)Croce and Hein]{Croce020a}
Francesco Croce and Matthias Hein.
\newblock Reliable evaluation of adversarial robustness with an ensemble of
  diverse parameter-free attacks.
\newblock In \emph{Proc. of ICML}, 2020.

\bibitem[{De Palma} et~al.(2022){De Palma}, Bunel, Dvijotham, Kumar, and
  Stanforth]{PalmaIBPR22}
Alessandro {De Palma}, Rudy Bunel, Krishnamurthy Dvijotham, M.~Pawan Kumar, and
  Robert Stanforth.
\newblock {IBP} regularization for verified adversarial robustness via
  branch-and-bound.
\newblock \emph{ArXiv preprint}, abs/2206.14772, 2022.

\bibitem[{De Palma} et~al.(2024){De Palma}, Bunel, Dvijotham, Kumar, Stanforth,
  and Lomuscio]{palma2024expressive}
Alessandro {De Palma}, Rudy~R Bunel, Krishnamurthy~Dj Dvijotham, M.~Pawan
  Kumar, Robert Stanforth, and Alessio Lomuscio.
\newblock Expressive losses for verified robustness via convex combinations.
\newblock In \emph{Proc. of ICLR}, 2024.

\bibitem[Duchi et~al.(2012)Duchi, Bartlett, and
  Wainwright]{duchi2012randomized}
John~C. Duchi, Peter~L. Bartlett, and Martin~J. Wainwright.
\newblock Randomized smoothing for stochastic optimization, 2012.

\bibitem[Ferrari et~al.(2022)Ferrari, M{\"{u}}ller, Jovanović, and
  Vechev]{FerrariMJV22}
Claudio Ferrari, Mark~Niklas M{\"{u}}ller, Nikola Jovanović, and Martin~T.
  Vechev.
\newblock Complete verification via multi-neuron relaxation guided
  branch-and-bound.
\newblock In \emph{Proc. of ICLR}, 2022.

\bibitem[Foret et~al.(2020)Foret, Kleiner, Mobahi, and
  Neyshabur]{foret2020sharpness}
Pierre Foret, Ariel Kleiner, Hossein Mobahi, and Behnam Neyshabur.
\newblock Sharpness-aware minimization for efficiently improving
  generalization.
\newblock \emph{arXiv preprint arXiv:2010.01412}, 2020.

\bibitem[Gehr et~al.(2018)Gehr, Mirman, Drachsler{-}Cohen, Tsankov, Chaudhuri,
  and Vechev]{GehrMDTCV18}
Timon Gehr, Matthew Mirman, Dana Drachsler{-}Cohen, Petar Tsankov, Swarat
  Chaudhuri, and Martin~T. Vechev.
\newblock {AI2:} safety and robustness certification of neural networks with
  abstract interpretation.
\newblock In \emph{Proc. of S\&P}, 2018.
\newblock \doi{10.1109/SP.2018.00058}.

\bibitem[Goodfellow et~al.(2015)Goodfellow, Shlens, and
  Szegedy]{GoodfellowSS14}
Ian~J. Goodfellow, Jonathon Shlens, and Christian Szegedy.
\newblock Explaining and harnessing adversarial examples.
\newblock In \emph{Proc. of ICLR}, 2015.

\bibitem[Gowal et~al.(2018)Gowal, Dvijotham, Stanforth, Bunel, Qin, Uesato,
  Arandjelovic, Mann, and Kohli]{GowalIBP2018}
Sven Gowal, Krishnamurthy Dvijotham, Robert Stanforth, Rudy Bunel, Chongli Qin,
  Jonathan Uesato, Relja Arandjelovic, Timothy~A. Mann, and Pushmeet Kohli.
\newblock On the effectiveness of interval bound propagation for training
  verifiably robust models.
\newblock \emph{ArXiv preprint}, abs/1810.12715, 2018.

\bibitem[He et~al.(2015)He, Zhang, Ren, and Sun]{HeZRS15}
Kaiming He, Xiangyu Zhang, Shaoqing Ren, and Jian Sun.
\newblock Delving deep into rectifiers: Surpassing human-level performance on
  imagenet classification.
\newblock In \emph{Proc. of ICCV}, 2015.
\newblock \doi{10.1109/ICCV.2015.123}.

\bibitem[Horv{\'{a}}th et~al.(2022)Horv{\'{a}}th, M{\"{u}}ller, Fischer, and
  Vechev]{HorvathM0V22}
Mikl{\'{o}}s~Z. Horv{\'{a}}th, Mark~Niklas M{\"{u}}ller, Marc Fischer, and
  Martin~T. Vechev.
\newblock Boosting randomized smoothing with variance reduced classifiers.
\newblock In \emph{{ICLR}}. OpenReview.net, 2022.

\bibitem[Jeong et~al.(2021)Jeong, Park, Kim, Lee, Kim, and Shin]{JeongPKLKS21}
Jongheon Jeong, Sejun Park, Minkyu Kim, Heung{-}Chang Lee, Do{-}Guk Kim, and
  Jinwoo Shin.
\newblock Smoothmix: Training confidence-calibrated smoothed classifiers for
  certified robustness.
\newblock In \emph{NeurIPS}, pp.\  30153--30168, 2021.

\bibitem[Jeong et~al.(2023)Jeong, Kim, and Shin]{JeongKS23}
Jongheon Jeong, Seojin Kim, and Jinwoo Shin.
\newblock Confidence-aware training of smoothed classifiers for certified
  robustness.
\newblock In \emph{{AAAI}}, pp.\  8005--8013. {AAAI} Press, 2023.

\bibitem[Jovanović et~al.(2022)Jovanović, Balunović, Baader, and
  Vechev]{JovanovicBBV22}
Nikola Jovanović, Mislav Balunović, Maximilian Baader, and Martin~T. Vechev.
\newblock On the paradox of certified training.
\newblock \emph{Trans. Mach. Learn. Res.}, 2022.

\bibitem[Katz et~al.(2017)Katz, Barrett, Dill, Julian, and
  Kochenderfer]{KatzBDJK17}
Guy Katz, Clark~W. Barrett, David~L. Dill, Kyle Julian, and Mykel~J.
  Kochenderfer.
\newblock Reluplex: An efficient {SMT} solver for verifying deep neural
  networks.
\newblock \emph{ArXiv preprint}, abs/1702.01135, 2017.

\bibitem[Kingma \& Ba(2015)Kingma and Ba]{KingmaB14}
Diederik~P. Kingma and Jimmy Ba.
\newblock Adam: {A} method for stochastic optimization.
\newblock In Yoshua Bengio and Yann LeCun (eds.), \emph{Proc. of ICLR}, 2015.

\bibitem[Krizhevsky et~al.(2009)Krizhevsky, Hinton,
  et~al.]{krizhevsky2009learning}
Alex Krizhevsky, Geoffrey Hinton, et~al.
\newblock Learning multiple layers of features from tiny images.
\newblock 2009.

\bibitem[Lange et~al.(2023)Lange, Tang, and Tian]{lange2023neuroevobench}
Robert~Tjarko Lange, Yujin Tang, and Yingtao Tian.
\newblock Neuroevobench: Benchmarking evolutionary optimizers for deep learning
  applications.
\newblock In \emph{Proc. of NeurIPS Datasets and Benchmarks Track}, 2023.

\bibitem[Le \& Yang(2015)Le and Yang]{Le2015TinyIV}
Ya~Le and Xuan~S. Yang.
\newblock Tiny imagenet visual recognition challenge.
\newblock \emph{CS 231N}, \penalty0 (7), 2015.

\bibitem[LeCun et~al.(2010)LeCun, Cortes, and Burges]{lecun2010mnist}
Yann LeCun, Corinna Cortes, and CJ~Burges.
\newblock Mnist handwritten digit database.
\newblock \emph{ATT Labs [Online]. Available:
  http://yann.lecun.com/exdb/mnist}, 2010.

\bibitem[Lee et~al.(2021)Lee, Lee, Park, and Lee]{LeeLPL21}
Sungyoon Lee, Woojin Lee, Jinseong Park, and Jaewook Lee.
\newblock Towards better understanding of training certifiably robust models
  against adversarial examples.
\newblock In \emph{NeurIPS}, pp.\  953--964, 2021.

\bibitem[Madry et~al.(2018)Madry, Makelov, Schmidt, Tsipras, and
  Vladu]{MadryMSTV18}
Aleksander Madry, Aleksandar Makelov, Ludwig Schmidt, Dimitris Tsipras, and
  Adrian Vladu.
\newblock Towards deep learning models resistant to adversarial attacks.
\newblock In \emph{Proc. of ICLR}, 2018.

\bibitem[Mao et~al.(2023{\natexlab{a}})Mao, Mueller, Fischer, and
  Vechev]{mao2023connecting}
Yuhao Mao, Mark~Niklas Mueller, Marc Fischer, and Martin Vechev.
\newblock Connecting certified and adversarial training.
\newblock In \emph{Proc. of NeurIPS}, 2023{\natexlab{a}}.

\bibitem[Mao et~al.(2023{\natexlab{b}})Mao, M{\"{u}}ller, Fischer, and
  Vechev]{MaoMFV24}
Yuhao Mao, Mark~Niklas M{\"{u}}ller, Marc Fischer, and Martin~T. Vechev.
\newblock Understanding certified training with interval bound propagation.
\newblock \emph{CoRR}, abs/2306.10426, 2023{\natexlab{b}}.
\newblock \doi{10.48550/ARXIV.2306.10426}.

\bibitem[Mao et~al.(2024)Mao, Balauca, and Vechev]{mao2024ctbench}
Yuhao Mao, Stefan Balauca, and Martin Vechev.
\newblock Ctbench: A library and benchmark for certified training.
\newblock \emph{arXiv preprint arXiv:2406.04848}, 2024.

\bibitem[Mao et~al.(2025)Mao, Zhang, and Vechev]{mao2025multineuron}
Yuhao Mao, Yani Zhang, and Martin Vechev.
\newblock On the expressiveness of multi-neuron convex relaxations, 2025.
\newblock URL \url{https://arxiv.org/abs/2410.06816}.

\bibitem[Mirman et~al.(2018)Mirman, Gehr, and Vechev]{MirmanGV18}
Matthew Mirman, Timon Gehr, and Martin~T. Vechev.
\newblock Differentiable abstract interpretation for provably robust neural
  networks.
\newblock In Jennifer~G. Dy and Andreas Krause (eds.), \emph{Proc. of ICML},
  2018.

\bibitem[Mirman et~al.(2022)Mirman, Baader, and Vechev]{MirmanBV22}
Matthew Mirman, Maximilian Baader, and Martin~T. Vechev.
\newblock The fundamental limits of neural networks for interval certified
  robustness.
\newblock \emph{Trans. Mach. Learn. Res.}, 2022.

\bibitem[M{\"{u}}ller et~al.(2023)M{\"{u}}ller, Eckert, Fischer, and
  Vechev]{MullerE0V23}
Mark~Niklas M{\"{u}}ller, Franziska Eckert, Marc Fischer, and Martin~T. Vechev.
\newblock Certified training: Small boxes are all you need.
\newblock In \emph{Proc. of ICLR}, 2023.

\bibitem[Owen et~al.(2019)Owen, Panken, Moss, Alvarez, and
  Leeper]{owen2019acas}
Michael~P Owen, Adam Panken, Robert Moss, Luis Alvarez, and Charles Leeper.
\newblock Acas xu: Integrated collision avoidance and detect and avoid
  capability for uas.
\newblock In \emph{2019 IEEE/AIAA 38th Digital Avionics Systems Conference
  (DASC)}, pp.\  1--10. IEEE, 2019.

\bibitem[Paszke et~al.(2019)Paszke, Gross, Massa, Lerer, Bradbury, Chanan,
  Killeen, Lin, Gimelshein, Antiga, Desmaison, K{\"{o}}pf, Yang, DeVito,
  Raison, Tejani, Chilamkurthy, Steiner, Fang, Bai, and
  Chintala]{PaszkeGMLBCKLGA19}
Adam Paszke, Sam Gross, Francisco Massa, Adam Lerer, James Bradbury, Gregory
  Chanan, Trevor Killeen, Zeming Lin, Natalia Gimelshein, Luca Antiga, Alban
  Desmaison, Andreas K{\"{o}}pf, Edward Yang, Zachary DeVito, Martin Raison,
  Alykhan Tejani, Sasank Chilamkurthy, Benoit Steiner, Lu~Fang, Junjie Bai, and
  Soumith Chintala.
\newblock Pytorch: An imperative style, high-performance deep learning library.
\newblock In \emph{Proc. of NeurIPS}, 2019.

\bibitem[Salman et~al.(2019)Salman, Li, Razenshteyn, Zhang, Zhang, Bubeck, and
  Yang]{SalmanLRZZBY19}
Hadi Salman, Jerry Li, Ilya~P. Razenshteyn, Pengchuan Zhang, Huan Zhang,
  S{\'{e}}bastien Bubeck, and Greg Yang.
\newblock Provably robust deep learning via adversarially trained smoothed
  classifiers.
\newblock In \emph{Proc. of NeurIPS}, 2019.

\bibitem[Salman et~al.(2020)Salman, Sun, Yang, Kapoor, and
  Kolter]{SalmanSYKK20}
Hadi Salman, Mingjie Sun, Greg Yang, Ashish Kapoor, and J.~Zico Kolter.
\newblock Denoised smoothing: {A} provable defense for pretrained classifiers.
\newblock In \emph{NeurIPS}, 2020.

\bibitem[Sehnke et~al.(2010)Sehnke, Osendorfer, R{\"{u}}ckstie{\ss}, Graves,
  Peters, and Schmidhuber]{SehnkeORGPS10}
Frank Sehnke, Christian Osendorfer, Thomas R{\"{u}}ckstie{\ss}, Alex Graves,
  Jan Peters, and J{\"{u}}rgen Schmidhuber.
\newblock Parameter-exploring policy gradients.
\newblock \emph{Neural Networks}, 2010.
\newblock \doi{10.1016/J.NEUNET.2009.12.004}.

\bibitem[Shi et~al.(2021)Shi, Wang, Zhang, Yi, and Hsieh]{ShiWZYH21}
Zhouxing Shi, Yihan Wang, Huan Zhang, Jinfeng Yi, and Cho{-}Jui Hsieh.
\newblock Fast certified robust training with short warmup.
\newblock In Marc'Aurelio Ranzato, Alina Beygelzimer, Yann~N. Dauphin, Percy
  Liang, and Jennifer~Wortman Vaughan (eds.), \emph{Proc. of NeurIPS}, 2021.

\bibitem[Shoeb et~al.(2009)Shoeb, Carlson, Panken, and
  Denison]{shoeb2009micropower}
Ali Shoeb, Dave Carlson, Eric Panken, and Timothy Denison.
\newblock A micropower support vector machine based seizure detection
  architecture for embedded medical devices.
\newblock In \emph{2009 Annual International Conference of the IEEE Engineering
  in Medicine and Biology Society}, pp.\  4202--4205. IEEE, 2009.

\bibitem[Singh et~al.(2018)Singh, Gehr, Mirman, P{\"{u}}schel, and
  Vechev]{SinghGMPV18}
Gagandeep Singh, Timon Gehr, Matthew Mirman, Markus P{\"{u}}schel, and
  Martin~T. Vechev.
\newblock Fast and effective robustness certification.
\newblock In \emph{Proc. of NeurIPS}, 2018.

\bibitem[Singh et~al.(2019)Singh, Gehr, P{\"{u}}schel, and Vechev]{SinghGPV19}
Gagandeep Singh, Timon Gehr, Markus P{\"{u}}schel, and Martin~T. Vechev.
\newblock An abstract domain for certifying neural networks.
\newblock \emph{Proc. of POPL}, 2019.
\newblock \doi{10.1145/3290354}.

\bibitem[Starnes et~al.(2023)Starnes, Dereventsov, and
  Webster]{starnes2023gaussian}
Andrew Starnes, Anton Dereventsov, and Clayton Webster.
\newblock Gaussian smoothing gradient descent for minimizing high-dimensional
  non-convex functions, 2023.

\bibitem[Sun et~al.(2024)Sun, Mao, M{\"{u}}ller, and Vechev]{abs-2410-06895}
Chenhao Sun, Yuhao Mao, Mark~Niklas M{\"{u}}ller, and Martin~T. Vechev.
\newblock Average certified radius is a poor metric for randomized smoothing.
\newblock \emph{CoRR}, abs/2410.06895, 2024.

\bibitem[Szegedy et~al.(2014)Szegedy, Zaremba, Sutskever, Bruna, Erhan,
  Goodfellow, and Fergus]{SzegedyZSBEGF13}
Christian Szegedy, Wojciech Zaremba, Ilya Sutskever, Joan Bruna, Dumitru Erhan,
  Ian~J. Goodfellow, and Rob Fergus.
\newblock Intriguing properties of neural networks.
\newblock In \emph{Proc. of ICLR}, 2014.

\bibitem[Toklu et~al.(2023)Toklu, Atkinson, Micka, Liskowski, and
  Srivastava]{toklu2023evotorch}
Nihat~Engin Toklu, Timothy Atkinson, Vojtěch Micka, Paweł Liskowski, and
  Rupesh~Kumar Srivastava.
\newblock Evotorch: Scalable evolutionary computation in python, 2023.

\bibitem[Tram{\`{e}}r et~al.(2020)Tram{\`{e}}r, Carlini, Brendel, and
  Madry]{TramerCBM20}
Florian Tram{\`{e}}r, Nicholas Carlini, Wieland Brendel, and Aleksander Madry.
\newblock On adaptive attacks to adversarial example defenses.
\newblock In \emph{Proc. of NeurIPS}, 2020.

\bibitem[Vaishnavi et~al.(2022)Vaishnavi, Eykholt, and Rahmati]{VaishnaviER22}
Pratik Vaishnavi, Kevin Eykholt, and Amir Rahmati.
\newblock Accelerating certified robustness training via knowledge transfer.
\newblock In \emph{NeurIPS}, 2022.

\bibitem[Wang et~al.(2018)Wang, Pei, Whitehouse, Yang, and Jana]{WangPWYJ18}
Shiqi Wang, Kexin Pei, Justin Whitehouse, Junfeng Yang, and Suman Jana.
\newblock Efficient formal safety analysis of neural networks.
\newblock In \emph{Proc. of NeurIPS}, 2018.

\bibitem[Wang et~al.(2021)Wang, Zhang, Xu, Lin, Jana, Hsieh, and
  Kolter]{WangZXLJHK21}
Shiqi Wang, Huan Zhang, Kaidi Xu, Xue Lin, Suman Jana, Cho{-}Jui Hsieh, and
  J.~Zico Kolter.
\newblock Beta-crown: Efficient bound propagation with per-neuron split
  constraints for neural network robustness verification.
\newblock In \emph{Proc. of NeurIPS}, 2021.

\bibitem[Weng et~al.(2018)Weng, Zhang, Chen, Song, Hsieh, Daniel, Boning, and
  Dhillon]{WengZCSHDBD18}
Tsui{-}Wei Weng, Huan Zhang, Hongge Chen, Zhao Song, Cho{-}Jui Hsieh, Luca
  Daniel, Duane~S. Boning, and Inderjit~S. Dhillon.
\newblock Towards fast computation of certified robustness for relu networks.
\newblock In \emph{Proc. of ICML}, 2018.

\bibitem[Wong \& Kolter(2018)Wong and Kolter]{WongK18}
Eric Wong and J.~Zico Kolter.
\newblock Provable defenses against adversarial examples via the convex outer
  adversarial polytope.
\newblock In \emph{Proc. of ICML}, 2018.

\bibitem[Wong et~al.(2018)Wong, Schmidt, Metzen, and Kolter]{WongSMK18}
Eric Wong, Frank~R. Schmidt, Jan~Hendrik Metzen, and J.~Zico Kolter.
\newblock Scaling provable adversarial defenses.
\newblock In \emph{Proc. of NeurIPS}, 2018.

\bibitem[Xu et~al.(2020)Xu, Zhang, Wang, Wang, Jana, Lin, and
  Hsieh]{xu2020fast}
Kaidi Xu, Huan Zhang, Shiqi Wang, Yihan Wang, Suman Jana, Xue Lin, and Cho-Jui
  Hsieh.
\newblock Fast and complete: Enabling complete neural network verification with
  rapid and massively parallel incomplete verifiers.
\newblock \emph{arXiv preprint arXiv:2011.13824}, 2020.

\bibitem[Xu et~al.(2021)Xu, Zhang, Wang, Wang, Jana, Lin, and
  Hsieh]{XuZ0WJLH21}
Kaidi Xu, Huan Zhang, Shiqi Wang, Yihan Wang, Suman Jana, Xue Lin, and
  Cho{-}Jui Hsieh.
\newblock Fast and complete: Enabling complete neural network verification with
  rapid and massively parallel incomplete verifiers.
\newblock In \emph{Proc. of ICLR}, 2021.

\bibitem[Zhang et~al.(2018)Zhang, Weng, Chen, Hsieh, and Daniel]{ZhangWCHD18}
Huan Zhang, Tsui{-}Wei Weng, Pin{-}Yu Chen, Cho{-}Jui Hsieh, and Luca Daniel.
\newblock Efficient neural network robustness certification with general
  activation functions.
\newblock In \emph{Proc. of NeurIPS}, 2018.

\bibitem[Zhang et~al.(2020)Zhang, Chen, Xiao, Gowal, Stanforth, Li, Boning, and
  Hsieh]{ZhangCXGSLBH20}
Huan Zhang, Hongge Chen, Chaowei Xiao, Sven Gowal, Robert Stanforth, Bo~Li,
  Duane~S. Boning, and Cho{-}Jui Hsieh.
\newblock Towards stable and efficient training of verifiably robust neural
  networks.
\newblock In \emph{Proc. of ICLR}, 2020.

\bibitem[Zhang et~al.(2022)Zhang, Wang, Xu, Li, Li, Jana, Hsieh, and
  Kolter]{ZhangWXLLJ22}
Huan Zhang, Shiqi Wang, Kaidi Xu, Linyi Li, Bo~Li, Suman Jana, Cho{-}Jui Hsieh,
  and J.~Zico Kolter.
\newblock General cutting planes for bound-propagation-based neural network
  verification.
\newblock \emph{ArXiv preprint}, abs/2208.05740, 2022.

\end{thebibliography}
\bibliographystyle{tmlr}

\ifincludeappendixx
	\clearpage
	\appendix
	\onecolumn
	
\section{Notation}
\label{sec:notation}

We use the following notation throughout this work:
\begin{itemize}[leftmargin=*]
    \item $\vx \in \bc{X} \subseteq \R^d$: Input vector to the neural network, assumed to lie in an input domain $\bc{X}$.
    \item $\bc{Y} = \{1, \dots, n\}$: Set of class labels for an $n$-class classification task.
    \item $\vf_{\bs{\theta}}(\vx) \in \R^n$: Neural network output (logits) for input $\vx$, parameterized by weights $\bs{\theta}$.
    \item $F(\vx) = \argmax_i \vf_{\bs{\theta}}(\vx)_i$: Induced classifier assigning input $\vx$ to the class with the highest logit.
    \item $\mathcal{B}^\epsilon_p(\vx) = \{\vx' \in \mathcal{X} \mid \| \vx - \vx' \|_p \leq \epsilon\}$: Closed $\ell_p$-ball of radius $\epsilon$ centered at $\vx$, representing the set of admissible perturbations.
    \item $\bc{L}_\text{CE}(\vy, t) = \ln\big(1 + \sum_{i \neq t} \exp(y_i - y_t)\big)$: Cross-entropy loss expressed in terms of the logit differences.
    \item $\vl, \vu$: Lower and upper bounds on intermediate or output activations, typically obtained via convex relaxations.
    \item $\mA_l, \vb_l$ and $\mA_u, \vb_u$: Coefficients defining affine lower and upper bounds on $\vf_{\bs{\theta}}(\vx)$ over the perturbation region.
    \item $\vy^\Delta := \vy - y_t$: Logit difference vector between logits $\vy$ and the target class $y_t$.
    \item $\overline{\vy}^\Delta$: Upper bound on the logit differences, used to compute the robust loss in certified training.
    \item $\vtheta \in \sR^d$: Parameter vector of the model.
    \item $L(\vtheta): \sR^d \rightarrow \sR$: Original (possibly non-smooth or discontinuous) loss function.
    \item $\mathcal{N}(\vzero, \sigma^2 \mI)$ or $\mathcal{N}(\vzero, \sigma^2)$: Isotropic Gaussian distribution with zero mean and standard deviation $\sigma$ (scalar).
    \item $\mathcal{N}(\vzero, \vsigma^2 \mI)$: Anisotropic Gaussian distribution with zero mean and standard deviation $\vsigma$ (vector).
    \item $\bs{\epsilon} \sim \mathcal{N}(\vzero, \sigma^2 \mI)$: Gaussian noise vector used for smoothing.
    \item $L_\sigma(\vtheta) := \E_{\bs{\epsilon} \sim \mathcal{N}(\vzero, \sigma^2 \mI)} L(\vtheta + \bs{\epsilon})$: Gaussian smoothed version of the loss.
    \item $D(f)$: Deviation from convexity of function $f$, defined as \\ $D(f) := \max_{\vx, \vy \in \R^d, \lambda \in [0,1]} \left[ f(\lambda \vx + (1-\lambda)\vy) - \lambda f(\vx) - (1-\lambda) f(\vy) \right]$.
    \item $L$-Lipschitz continuity: A function $f$ is Lipschitz continuous if $\exists L > 0$ s.t.\ $|f(\vx) - f(\vy)| \leq L\|\vx - \vy\|$ for all $\vx, \vy$.
    \item $\relu(\cdot)$: Rectified Linear Unit activation function.
    \item $\mathds{1}_{w > 0}$: Indicator function that is 1 if $w > 0$ and 0 otherwise.
    \item $\frac{f(x+h) - f(x)}{h}$: Finite difference approximation used to estimate discontinuity magnitude.
    \item $\frac{f(x-h)-2f(x)+f(x+h)}{h^2}$: Second-order finite difference used to estimate curvature (non-smoothness).
    \item $\sigma$, $\vsigma_{PGPE}$, $\sigma_{RGS}$: Standard deviation of Gaussian used in GLS. A vector in the case of PGPE, a scalar hyperparameter for RGS.
    \item $n_{ps}$: Population size used in sampling-based GLS estimators (e.g., PGPE, RGS).
\end{itemize}

Unless otherwise noted, we focus on robustness with respect to the $\ell_\infty$ norm, i.e., $\mathcal{B}^\epsilon(\vx) = \{\vx' \in \mathcal{X} \mid \|\vx - \vx'\|_\infty \leq \epsilon\}$.

\section{Theoretical Power of GLS} \label{sec:proofs}

\subsection{Proofs}
\label{app:proofs}

Throughout our proof, we will denote the probability density function of the Gaussian distribution $\gN(\vzero, \sigma^2 \mI)$ as $p_{\sigma}(x)$. A ball of radius $r$ is defined as $B(r) := \{\vx \mid \|\vx\|\le r\}$. Without explicit mention, the norms are $L_2$ norm.

\begin{lemma} \label{lem:convolution}
    Assume the existence of the smoothed loss function. Gaussian Loss Smoothing is equivalent to performing a convolution of the loss function with a Gaussian kernel, that is, $L_\sigma(\vtheta) = \left[L * \gN(\vzero, \sigma^2 \mI)\right](\vtheta)$.
\end{lemma}
\begin{proof}
    Remember that $L_\sigma(\vtheta) := \E_{\bs{\epsilon} \sim \gN(\vzero, \sigma^{2} \mI)} L(\vtheta+\bs{\epsilon})$. Let $\vx := \vtheta + \bs{\epsilon}$ and use $p_\sigma(\bs{\epsilon}) = p_\sigma(-\bs{\epsilon})$ due to symmetry, we have
    \begin{align} \label{eq:convolution}
        L_\sigma(\vtheta) &= \E_{\bs{\epsilon} \sim \gN(\vzero, \sigma^{2} \mI)} L(\vtheta+\bs{\epsilon}) \nonumber\\
        &= \int_{\R^d} L(\vtheta + \bs{\epsilon}) p_\sigma(\bs{\epsilon}) d\bs{\epsilon} \nonumber\\
        &= \int_{\R^d} L(\vtheta + \bs{\epsilon}) p_\sigma(-\bs{\epsilon}) d\bs{\epsilon} \nonumber\\
        &= \int_{\R^d} p_\sigma(\vtheta - \vx) L(\vx) d \vx \\
        &= \left[L * \gN(\vzero, \sigma^2 \mI)\right](\vtheta). \nonumber
    \end{align}
\end{proof}

\begin{proposition} \label{prop:smooth}
    Assume the nonnegative loss function $L(\vtheta): \sR^d \rightarrow \sR$ have bounded growth, that is, $L(\vtheta) \exp(-\|\vtheta\|^{2-\delta}) \le M$ for some $\delta<2$ and $M>0$. Then, $L_\sigma$ exists and is infinitely differentiable.
\end{proposition}
\begin{proof}
    We prove existence of $L_\sigma$ first. \cref{eq:convolution} shows that this is equivalent to prove the convergence of the integral. Given a fixed $\vtheta$, $p_\sigma(\vtheta - \vx) \propto \exp(-\frac{1}{2\sigma^2} \|\vx - \vtheta\|^2)$, thus $\exists \alpha_1, \beta_1, M_1>0$, such that $p_\sigma(\vtheta - \vx) \le \alpha_1 \exp(-\beta_1 \|\vx\|^2)$ when $\|\vx\|>M_1$. Therefore, 
    \begin{align*}
        & \quad \int_{\R^d \setminus B(M_1)} p_\sigma(\vtheta - \vx) L(\vx) d \vx \\
        &\le \int_{\R^d \setminus B(M_1)} \alpha_1 \exp(-\beta_1 \|\vx\|^2) L(\vx) d \vx \\
        &\le \int_{\R^d \setminus B(M_1)} \alpha_1 \exp(-\beta_1 \|\vx\|^2) M \exp(\|\vx\|^{2-\delta}) d \vx \\
        &\le \alpha_1 M \int_{\R^d \setminus B(M_1)} \exp(-\beta_1 \|\vx\|^2 + \|\vx\|^{2-\delta}) d \vx.
    \end{align*}
    Further, $\exists \beta_2>0, M_2 \ge M_1$, such that $\exp(-\beta_1 \|\vx\|^2 + \|\vx\|^{2-\delta}) \le \exp(-\beta_2 \|\vx\|^{\delta/2})$ when $\|\vx\|>M_2$. Therefore,
    \begin{align*}
        & \quad \int_{\R^d \setminus B(M_2)} p_\sigma(\vtheta - \vx) L(\vx) d \vx \\
        &\le \alpha_1 M \int_{\R^d \setminus B(M_2)} \exp(-\beta_1 \|\vx\|^2 + \|\vx\|^{2-\delta}) d \vx \\
        &\le \alpha_1 M \left[\int_{\R^d \setminus B(M_2)} \exp(-\beta_2 \|\vx\|^{\delta/2}) d \vx \right]. \\
    \end{align*}
    Note that $\exp(-\beta_2 \|\vx\|^{\delta/2})$ decays faster than $\frac{1}{\|\vx\|^2}$ and $\int_{\R^d \setminus B(M_2)} \frac{1}{\|\vx\|^2} d\vx$ is bounded. Thus, $\forall \epsilon>0$, $\exists M_3 \ge M_2$, such that $\int_{\R^d \setminus B(M_3)} p_\sigma(\vtheta - \vx) L(\vx) d \vx < \epsilon$. Therefore, $L_\sigma$ exists.

    Now we turn to its derivative. Using \cref{lem:convolution} and $(f * g)^\prime (t) = (f * g^\prime)(t)$, we know that any $n$-th (partial) derivative of $L_\sigma$ is $L * \frac{\partial^{(n)} p_\sigma}{\partial^{(n)} \vx}$, where $\partial^{(n)} \vx$ is a shorthand for the related variables. Since $n$-th partial derivative of a Gaussian pdf is a polynomial (Hermite polynomials) times a Gaussian pdf, we can bound it similarly to what we have done before, as we can still bound $\frac{\partial^{(n)} p_\sigma}{\partial^{(n)} \vx}$ with $\alpha_1 \exp(-\beta_1 \|\vx\|^2)$ under appropriate $\alpha_1, \beta_1, M_1$. Therefore, $L_\sigma$ is infinitely differentiable.
\end{proof}

\begin{lemma} \label{lem:Lipschitz}
    If $f$ is continuously differentiable, then $f$ is Lipschitz continuous within a compact set.
\end{lemma}
\begin{proof}
    Since $f$ has continuous first-order derivative, it suffices to show that the first-order gradients are bounded. This is trivial as a continuous function is bounded within any compact set.
\end{proof}

\begin{proposition} \label{prop:deviation_convexity}
    Assume $f*g$ exists, where $g$ is a probability density function. Then, the deviation from convexity of $f*g$ is smaller than or equal to the deviation from convexity of $f$, that is, $D(f*g) \le D(f)$. Equality holds iff $f$ is an affine function.
\end{proposition}
\begin{proof}
    \begin{align*}
        \delta[f*g; \vx, \vy, \lambda] &= f*g(\lambda \vx + (1-\lambda) \vy) - \lambda f*g(\vx) - (1-\lambda) f*g(\vy) \\
        &= \int_{\R^d} \left[f(\lambda \vx + (1-\lambda) \vy - \vz)  - \lambda f(\vx - \vz)  - (1-\lambda)  f(\vy - \vz) \right]g(\vz) d\vz \\
        &= \int_{\R^d} \delta[f; \vx-z, \vy-z, \lambda] g(\vz) d\vz \\
        &\le \max_{\vx,\vy \in \R^d; \lambda \in [0,1]} \delta[f; \vx, \vy, \lambda] \int_{\R^d}  g(\vz) d\vz \\
        &= D(f) \int_{\R^d}  g(\vz) d\vz \\
        &= D(f),
    \end{align*}
    where we used the fact that $g$ is a probability density function, thus $\int_{\R^d}  g(\vz) d\vz = 1$. The above equality holds iff $\delta[f; \vx, \vy, \lambda]$ is a constant function. Therefore, $D(f*g) = \max_{\vx,\vy \in \R^d; \lambda \in [0,1]} \delta[f*g; \vx, \vy, \lambda] \le D(f)$. Note that to take equality, it is necessary that $\delta[f*g; \vx, \vy, \lambda] = D(f)$ for some $\vx, \vy, \lambda$, thus $\delta[f; \vx, \vy, \lambda]$ still has to be a constant function. On the other hand, if $\delta[f; \vx, \vy, \lambda]$ is a constant function, then $\delta[f*g; \vx, \vy, \lambda] = D(f)$ for all $\vx, \vy, \lambda$, thus $D(f*g) = D(f)$. Therefore, $D(f*g) = D(f)$ iff $\delta[f; \vx, \vy, \lambda]$ is a constant function.

    Now we show that $\delta[f; \vx, \vy, \lambda]$ is a constant function iff $f$ is an affine function. If $f$ is an affine function, then $\delta[f; \vx, \vy, \lambda] = f(\lambda \vx + (1-\lambda) \vy) - \lambda f(\vx) - (1-\lambda) f(\vy) = 0$, thus $\delta[f; \vx, \vy, \lambda]$ is a constant function. On the other hand, if $\delta[f; \vx, \vy, \lambda]$ is a constant function, then $\exists C$ such that $\delta[f; \vx, \vy, \lambda] = C$ for all $\vx, \vy, \lambda$. Let $\vx = \vy = \vzero$, then $C = f(\vzero) - f(\vzero) = 0$, thus $f(\lambda \vx + (1-\lambda) \vy) = \lambda f(\vx) + (1-\lambda) f(\vy)$ for all $\vx, \vy, \lambda$. This means $f$ is an affine function.
\end{proof}

\glsinfdiff*
\begin{proof}
    By \cref{prop:smooth}, $L_\sigma$ exists and is infinitely differentiable. Further, by \cref{lem:convolution} and \cref{prop:deviation_convexity}, $D(L_\sigma) \le D(L)$; equality holds iff $L$ is an affine function. Assuming $\theta$ is in a compact set, by \cref{lem:Lipschitz}, $L_\sigma$ is Lipschitz continuous.
\end{proof}

\subsection{Alignment of Local and Global Minima under Gaussian Loss Smoothing} \label{app:alignment}

Without loss of generality, we consider a quantized function $f(x) = \sum_{i=0}^n a_i I(x \in [b_i, b_{i+1}])$, where $I$ is the threshold function and $-\infty = b_0 \le b_1 \le \dots \le b_n \le b_{n+1}=+\infty$. The global minimum of this function is $\min_i a_i$, achieved by $x \in [b_{i^*}, b_{i^*+1}]$ where $i^* \in \arg \min_i a_i$. Now, the derivative of its Gaussian smoothed loss is $f^\prime_\sigma(x) = \frac{1}{\sigma} \sum_{i=1}^n (a_i - a_{i-1}) p(\frac{b_i -x}{\sigma})$, where $p$ is the p.d.f. of the standard normal distribution.
One may immediately find that the minimum of the smoothed loss is scale-invariant: the minimum of $f_{c\sigma} (cx)$ with $b_i$ scaled by $c$ is the same as the minimum of $f_\sigma(x)$. Therefore, if we increase $\sigma$ to smoothen a fixed function, shallower minima with smaller widths will be smoothed out one by one. Taking $\sigma$ to $\infty$, we find that the derivative converges to zero, making the smoothed loss a constant function.

We use the same quantized function to study the effect smoothing has on the alignment of minimum points. As observed before, when we take $\sigma$ to $\infty$, the derivative on the whole domain converges to zero, so every point becomes a minimum, therefore we fail to get a proper alignment. On the other hand, by taking $\sigma$ to zero, the factor $p(\frac{b_i -x}{\sigma})$ becomes a Dirac delta function $\delta(x = b_i)$, thus every point except the boundary points becomes a local minimum, and we get the alignment of global minima.
Based on these intuitions, one can pick a $\sigma$ such that narrow local minima get smoothed out, and wide local minima are left close to their original locations, thus the optimization process can be guided towards the global minimum.

\subsection{Properties of Randomized Gradient Smoothing}\label{sec:rgs-properties}

\paragraph{Discontinuity} Considering again the quantized function defined in \cref{app:alignment}, we observe that the derivative of the original function is zero almost everywhere, so the smoothed gradient estimated by RGS will also be zero. This means that RGS may fail to find the minimum of certain discontinuous functions in general. However, in practice we rarely work with quantized loss functions we used for the analysis; instead, we can model the discontinuous loss function as $h(x) = f(x) + g(x)$, where $f(x)$ is discontinuous like the quantized function and $g(x)$ is continuous. In this case, the derivative of $h$ is equal to the derivative of $g$ almost everywhere, and thus the RGS algorithm will converge to the same locations when optimizing $h$ as when optimizing $g$. If the minima of $g$ and $h$ are sufficiently aligned, we can expect RGS to find a good minimum of $h$.

\paragraph{Higher Dimensions} In higher dimensions, however, the behavior of RGS becomes unpredictable, as not every discontinuous function $h$ can be decomposed into a continuous function $g$ and a quantized function $f$ (e.g. $h(x_1,x_2)=x_1\cdot \sign{(x_2)}$ consists of two plane sections separated by a discontinuity along the $x_1$-axis). In this case, the equivalent loss landscape that the RGS algorithm is optimizing is strongly dependent on the optimization path and the starting point and therefore cannot be defined.

\section{Ablation Studies}
\label{sec:ablations}

\begin{figure}[t]
	\centering
	\vspace{-8mm}
	\includegraphics[width=.6\linewidth]{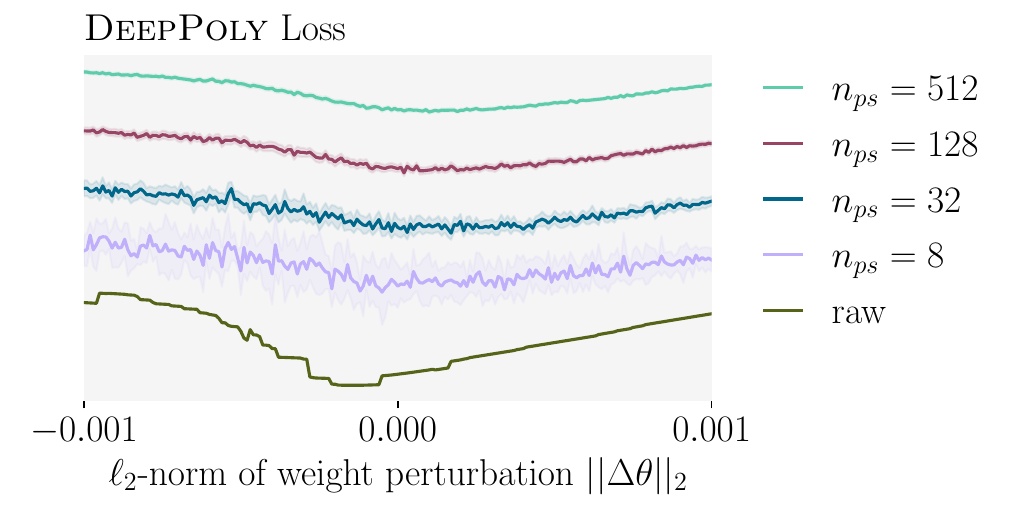}
	\vspace{-4mm}
	\caption{Effect of the population size $n_{ps}$ on the smoothness of the induced loss surface in \pgpe. Note that the 5 plots have been spaced by artificially adding offsets on the y-axis. This should not be regarded as a quantitative plot ordering the magnitude of the loss, but rather as a qualitative comparison of the smoothness induced by sampling with different population sizes.} \label{fig:popsize-ablation}
	\vspace{-5mm}
\end{figure}

\subsection{Population Size}
While \pgpe recovers Gaussian Loss Smoothing in expectation, the quality of the gradient approximation depends strongly on the population size $n_{ps}$.
In particular, a small population size $n_{ps}$ induces a high-variance estimate of the true smoothed loss, leading to noisy gradient estimates and thus slow learning or even stability issues. We illustrate this in \cref{fig:popsize-ablation} where we show the loss surface along the gradient direction for different population sizes. We observe that for small population sizes the loss surface is indeed very noisy, only becoming visually smooth at $n_{ps} = 512$. 
Additionally, \pgpe computes a gradient approximation in an $\tfrac{n_{ps}}{2}$-dimensional subspace, thus further increasing gradient variance if $n_{ps}$ is (too) small compared to the number of network parameters.

\begin{minipage}{\textwidth}
    \begin{minipage}[c]{0.5\textwidth}
            \centering
            \vspace{-3mm}
            \captionof{table}{Effect of the population size $n_{ps}$ on accuracy and training time with \pgpe + \deeppoly training on \cnnt.}
            \label{tab:popsize-ablation}
            \vspace{1mm}
            \centering
            \small
            \begin{tabular}{@{}cccc}
                \toprule
                {Popsize} & {Nat. [\%]} & {Cert. [\%]} & {GPU h} \\
                \midrule
                Init & 97.14 & 94.02 & - \\
                64   & 97.22 & 94.07 & 88 \\
                128  & 97.22 & 94.13 & 160 \\
                256  & 97.30 & 94.19 & 304 \\
                512  & 97.27 & 94.22 & 596 \\
                1024 & 97.43 & 94.50 & 1192  \\
               
                \bottomrule
            \end{tabular}
            \vspace{-1.0em}
    \end{minipage}
    \hfill
    \begin{minipage}[c]{0.46\textwidth}
        \centering
            \includegraphics[width=.95\linewidth]{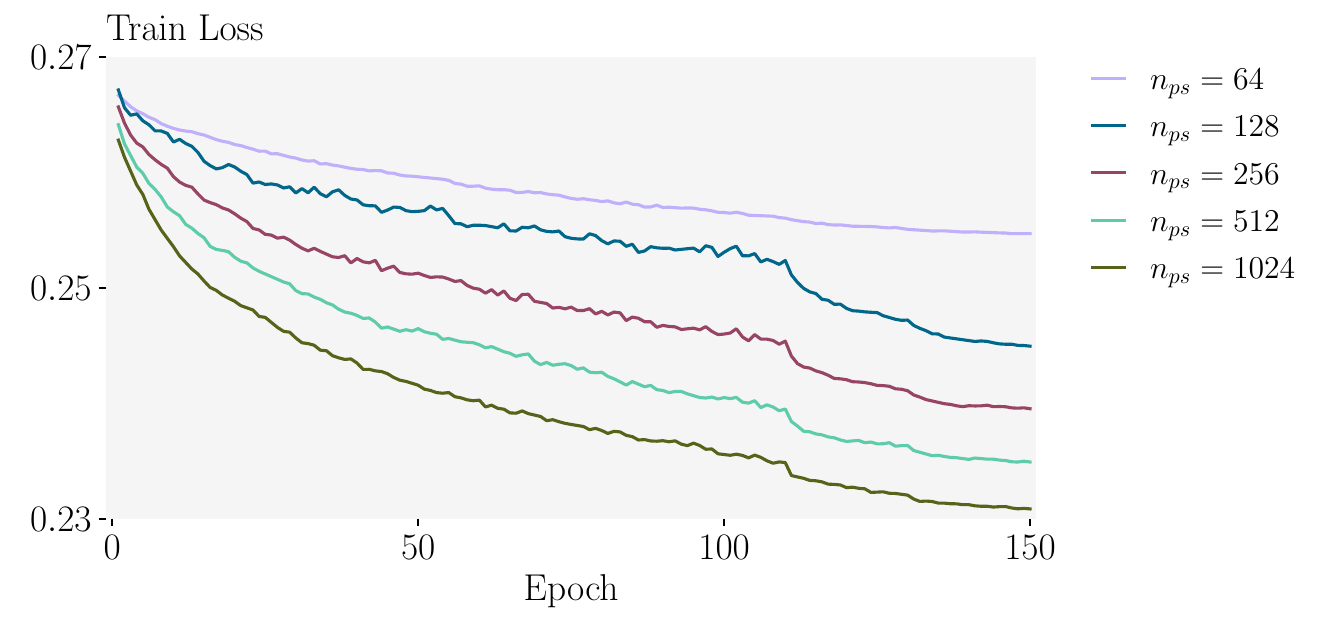}
            \vspace{-1mm}
            \captionof{figure}{Evolution of Train Loss during training with different values for popsize $n_{ps}$. Note that for $n_{ps}=64$ we trained with a lower learning rate because the value used in the other settings would make training unstable. } 
            \label{fig:train_dynamics}
    \end{minipage}
\end{minipage}

To assess the effect this has on the performance of \pgpe training, we train the same \cnnt on \mnist using population sizes between $64$ and $1024$, presenting results in \cref{tab:popsize-ablation}.
We observe that performance does indeed improve significantly with increasing population sizes (note the relative performance compared to initialization). This becomes even more pronounced when considering the training dynamics (see \cref{fig:train_dynamics}).
Unfortunately, the computational cost of \pgpe is significant and scales linearly in the population size. We thus choose $n_{ps}=256$ for all of our main experiments, as this already leads to training times of more than $1$ week on $8$ L4 GPUs for some experiments.

\paragraph{Train Dynamics when varying population size}
In \cref{fig:train_dynamics} we present the evolution of the Training Loss during training with different values for popsize $n_{ps}$. We observe significantly slower training as we decrease $n_{ps}$, confirming the theoretical prediction that using lower popsize decreases the quality of gradient estimations due to increased variance in the loss-sampling process.

\vspace{-2mm}
\subsection{Standard Deviation}
The standard deviation $\sigma$ used for Gaussian Loss Smoothing has a significant impact on the resulting loss surface as we illustrated in \cref{fig:empirical_smooth} and discussed in \cref{sec:gausian_loss_smoothing}. If $\sigma$ is chosen too small, the loss surface will still exhibit high sensitivity and gradients will only be meaningful very locally as discontinuities are barely smoothed. On the other hand, if $\sigma$ is chosen too large, the loss surface will become very flat and uninformative, preventing us from finding good solutions. 

When estimating the smoothed loss in \pgpe via sampling at moderate population sizes $n_{ps}$, the standard deviation $\sigma_{\pgpe}$ additionally affects the variance of the loss and thus gradient estimate. We illustrate this in \cref{fig:stdev-ablation}, where we not only see the increasing large-scale smoothing effect discussed above but also an increasing level of small-scale noise induced by a large $\sigma_{\pgpe}$ relative to the chosen population sizes $n_{ps}$.

To assess the effect this practically has on \pgpe training, we train for 50 epochs with different standard deviations $\sigma_{\pgpe}$ and present the results in \cref{fig:stdev_train}. As expected, we clearly observe that both too small and too large standard deviations lead to poor performance. However, and perhaps surprisingly, we find that training performance is relatively insensitive to the exact standard deviation as long as we are in the right order of magnitude between $10^-3$ and $10^-2$.

\vspace{2mm}
\begin{minipage}{\textwidth}
    \begin{minipage}[c]{0.52\textwidth}
        \centering
	\includegraphics[width=.9\linewidth]{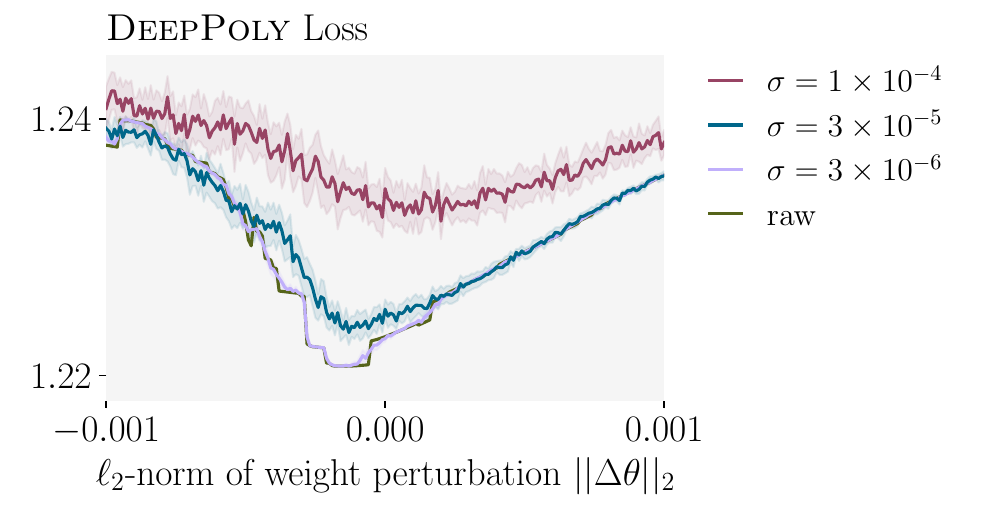}
	\vspace{-3mm}
	\captionof{figure}{Effect of the standard deviation $\sigma_\pgpe$ on the induced loss surface in \pgpe at a small population sizes of $n_{ps}=32$.} 
	\label{fig:stdev-ablation}%
    \end{minipage}
    \hfill
    \begin{minipage}[c]{0.44\textwidth}
        \centering
        \includegraphics[width=.9\linewidth]{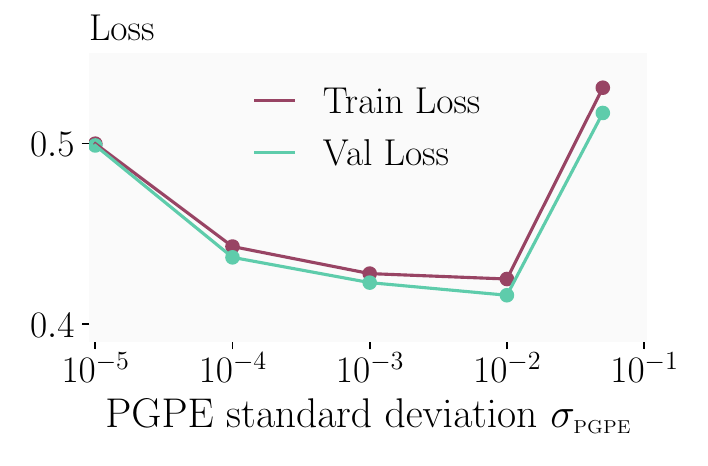}
        \vspace{-3mm}
        \captionof{figure}{Train and Validation Losses after 50 epochs of training for different values of $\sigma_{\pgpe}$.} %
        \vspace{-2mm}
        \label{fig:stdev_train}
    \end{minipage}
\end{minipage}

\subsection{Train Dynamics Comparison}
\label{sec:train-dynamics-comparison}

\begin{figure}[ht]
	\centering
	\includegraphics[width=.6\linewidth]{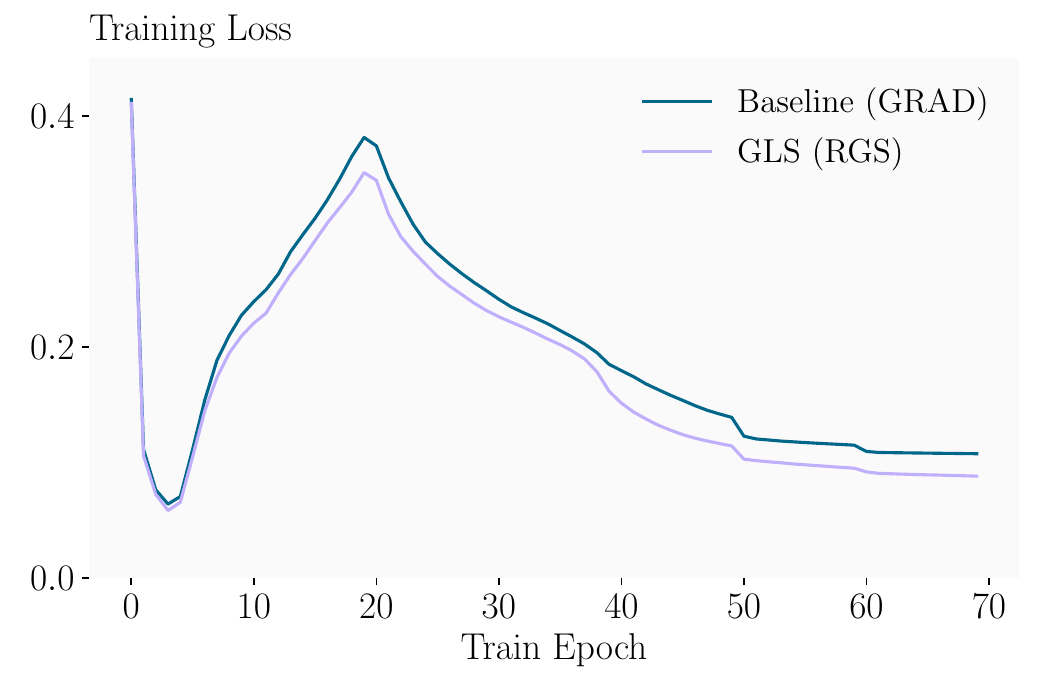}
	\vspace{-3mm}
	\caption{Evolution of training loss when training a \cnnf on \mnist $\epsilon=0.1$ using the \deeppoly convex relaxation with standard backpropagation (GRAD) and with GLS-based RGS.} %
	\vspace{-2mm}
	\label{fig:train-dynamics-comparison}
\end{figure}

In \cref{fig:train-dynamics-comparison} we show the differences between the training dynamics of GRAD-DeepPoly (Baseline) and RGS-DeepPoly for our CNN5 models trained on MNIST $\epsilon=0.1$. We can observe that using RGS we obtain lower losses even from the $\epsilon$-annealing phase (first 20 epochs), resulting in a better training quality and better model after training is done.

\section{Additional Experimental Data}

\subsection{Training CROWN-IBP-RGS on \cnns}
\label{app:crownibp-cnn7}

While \deeppoly-RGS is too computationally extensive for scaling to \cnns, we can use RGS in combination with \crownibp to prove that the advantages of GLS scale even to SOTA architectures. We present the results of training \cnns with \crownibp-RGS in \cref{tab:cnn7-results}. We observe that RGS significantly increases the performance of \crownibp when applied on \cnns without BatchNorm layers, but the improvement is less pronounced when using BatchNorm layers. 

In order to accomodate the use of BatchNorm layers with RGS, we compute the estimated gradients by using the BatchNorm statistics independently for each sample of perturbed weights. To obtain the test statistics, we reset the running stats to the population statistics for the mean trained network after each epoch, following the advice of \citet{mao2024ctbench}. While this approach is the most straightforward, it might not be the most effective way to use BatchNorm layers with RGS, and we leave the exploration of more sophisticated methods for future work.

\begin{table}[ht]
    \centering
    \caption{Comparison between \cnns networks trained with \crownibp-RGS and SOTA GRAD methods on small perturbation settings}
    \label{tab:cnn7-results}
    \centering
    \vspace{-1mm}
    \footnotesize
    \resizebox{0.85\linewidth}{!}{
		\begin{tabular}{cclccc}
            \toprule
            {Dataset} & \makecell{Network}  & Method  & \makecell{Nat. Acc. \\ $\text{[\%]}$}    & \makecell{Cert. Acc. \\ $\text{[\%]}$} & \makecell{Adv. Acc. \\ $\text{[\%]}$}  \\
            \midrule
            \multirow{7.5}{*}{\makecell{\mnist \\ $\epsilon_\infty=0.1$}}     
            & \multirow{4}{*}{\makecell{\cnns \\ with BN}}   
              & IBP           & 98.87 & 98.26 & 98.27 \\
            & & CROWN-IBP     & 99.10 & 98.13 & 98.22 \\
            & & CROWN-IBP-RGS & 99.11 & 98.05 & 98.09 \\
            & & TAPS          & 99.16 & 98.52 & 98.58 \\
            \cmidrule(rl){2-6}
            & \multirow{3}{*}{\makecell{\cnns \\ no BN}}   
              & IBP           & 98.50 & 97.40 & 97.42 \\
            & & CROWN-IBP     & 98.83 & 97.94 & 97.94 \\
            & & CROWN-IBP-RGS & 99.19 & 98.09 & 98.18 \\
            \midrule
            \multirow{7.5}{*}{\makecell{\cifar \\ $\epsilon_\infty=2/255$}}      
            & \multirow{4}{*}{\makecell{\cnns \\ with BN}}   
              & IBP           & 67.49 & 55.99 & 56.10 \\
            & & CROWN-IBP     & 70.90 & 58.80 & 59.93 \\
            & & CROWN-IBP-RGS & 70.82 & 59.04 & 60.19 \\
            & & MTL-IBP       & 78.82 & 64.41 & 67.69 \\
            \cmidrule(rl){2-6}
            & \multirow{3}{*}{\makecell{\cnns \\ no BN}}   
              & IBP           & 63.13 & 52.09 & 52.27 \\
            & & CROWN-IBP     & 67.82 & 55.36 & 56.62 \\
            & & CROWN-IBP-RGS & 68.46 & 56.30 & 57.37 \\
            \bottomrule
        \end{tabular}
    }
    \vspace{-4mm}
\end{table}

Finally, the results showcase that the promise of using GLS for certified training with tighter relaxations also scales to SOTA architectures.

\subsection{Comparison between RGS and PGPE on \cnnt and \cnnty}
\label{app:pgpe-rgs-comparison}

In \cref{tab:pgpe-rgs-comparison} we provide additional experimental data comparing the performance of PGPE and RGS on \cnnt and \cnnty. In the case of \cnnt, we observe that RGS and PGPE obtain similar performance on MNIST 0.1, but RGS is actually significantly better on CIFAR 2/255 (note that both are better than \ibp and \deeppoly trained with Adam). We hypothesize that this might be due to: (1) on \cifar the \cnnt network has more parameters than on \mnist (~5k vs ~7k due to different input sizes), thus the parameter space is larger; (2) we train for the same number of epochs on both datasets with \pgpe, but standard certified training has shown networks on \cifar to converge slower than on \mnist. As a result, since \pgpe trains slower due to the low-rank gradient problem, \cifar makes this worse, and insufficient training outweighs the theoretical benefit. These claims are further supported by our results for \cnnty trained on \mnist where we observe that \pgpe is significantly better than RGS, likely due to the smaller parameter space and faster convergence of the training.

\begin{table}[h]
    \centering
    \caption{Comparison of PGPE and RGS on \texttt{CNN3}}
    \label{tab:pgpe-rgs-comparison}
    \footnotesize
    \begin{tabular}{cclccc}
        \toprule
        Dataset & Network (params.)      & Method   & Nat. [\%] & Cert. [\%] & Adv. [\%]  \\
        \midrule
        \multirow{4}{*}{\makecell{\mnist \\ $\epsilon_\infty=0.1$}} 
        & \multirow{4}{*}{\cnnty (1.1k)}  & IBP-GRAD & 89.76 & 82.46 & 82.48 \\
                                       &  & \deeppoly-GRAD  & 89.24 & 68.47 & 68.57 \\
            &  & \deeppoly-PGPE  & \textbf{91.94} & \textbf{85.00} & \textbf{85.04} \\
                                       &  & \deeppoly-RGS   & 91.33 & 82.66 & 82.68 \\
        \midrule
        \multirow{4}{*}{\makecell{\mnist \\ $\epsilon_\infty=0.1$}} 
        & \multirow{4}{*}{\cnnt (5.2k)}  & IBP-GRAD & 96.02 & 91.23 & 91.23 \\
                                      &  & \deeppoly-GRAD  & 95.95 & 90.04 & 90.08 \\
                                      &  & \deeppoly-PGPE  & \textbf{97.44} & 91.53 & 91.79 \\
                                      &  & \deeppoly-RGS   & 97.37 & \textbf{91.88} & \textbf{92.03} \\
        \midrule
        \multirow{4}{*}{\makecell{\cifar\\ $\epsilon_\infty=2/255$}}
        & \multirow{4}{*}{\cnnt (6.8k)}  & IBP-GRAD & 48.05 & 37.69 & 37.70 \\
                                      &  & \deeppoly-GRAD  & 47.70 & 36.72 & 36.72 \\
                                      &  & \deeppoly-PGPE  & 54.17 & 38.95 & 40.20 \\
           &  & \deeppoly-RGS   & \textbf{54.93} & \textbf{41.14} & \textbf{42.03} \\
        \bottomrule
        \end{tabular}
\end{table}

\subsection{Comparison of GLS and Sharpness-Aware Minimization (SAM)}

The gradient computation by perturbing the network weights used in \pgpe and \rgs has some similarities with the Sharpness-Aware Minimization (SAM, \citet{foret2020sharpness}) algorithm. However, the SAM algotithm is fundamentally different to GLS. This is because GLS takes the expectation of neighborhood loss rather than the worst case loss; in fact, SAM is closer to adversarial training with FGSM \citep{GoodfellowSS14} rather than GLS. In particular, SAM does not resolve the discontinuity problem, while GLS provably solves it (\cref{thm:gls}). To see this, consider the threshold function $I(x>0)$ and an initial $x_0=0.1$. Any single-point gradient based methods (including SAM) will only get zero gradient, and thus cannot optimize it. Therefore, while it is likely that GLS has the benefit of reduced sharpness as well, GLS enjoys fundamentally different benefits to SAM.

To confirm this empirically, we apply SAM to IBP and \deeppoly training. Specifically, we update the parameters with gradients computed based on the adversarially perturbed network $w^\prime = w + \rho \times \nabla_w L / \|\nabla_w L\|_2$. We train with IBP and \deeppoly on MNIST $\epsilon=0.1$ with the same CNN3 architecture used in the paper. The results are shown in \cref{tab:sam-comparison}, all networks certified with MN-BaB.

\begin{table}[h]
    \caption{Comparison of GLS methods with SAM. \cnnt networks trained on \mnist $\epsilon=0.1$.}
    \label{tab:sam-comparison}
    \centering
    \begin{tabular}{lcc}
    \toprule
    Method               & Nat. [\%] & Cert. [\%] \\
    \midrule
    IBP-GRAD             & 96.02 & 91.23 \\
    IBP-SAM $\rho=0.1$   & 96.08 & 90.20 \\
    IBP-SAM $\rho=0.01$  & 96.32 & 93.32 \\
    IBP-SAM $\rho=0.001$ & 95.80 & 91.73 \\
    \midrule
    \deeppoly-GRAD              & 95.95 & 90.04 \\
    \deeppoly-SAM $\rho=0.1$    & 94.22 & 88.39 \\
    \deeppoly-SAM $\rho=0.01$   & 96.93 & 92.34 \\
    \deeppoly-SAM $\rho=0.001$  & 96.95 & 90.91 \\
    \deeppoly-PGPE              & 97.44 & 91.53 \\
    \deeppoly-RGS               & 97.37 & 91.88 \\
    \bottomrule
    \end{tabular}
\end{table}

We observe that for a correctly chosen hyperparameter ($\rho = 0.01$), SAM does indeed improve performance for IBP and DP. While SAM performs better than PGPE for this very shallow network, as expected from our previous theoretical analysis, it does not address the paradox. In particular, IBP-SAM still performs better than DP-SAM uniformly for every choice of $\rho$. While combining SAM with PGPE or other certified training methods might thus constitute an interesting future direction, it does not explain the reranking of approximation methods (\deeppoly-PGPE > IBP-PGPE vs IBP-SAM > \deeppoly-SAM) we observe for PGPE. We therefore conclude that the sharpness aware aspect of PGPE is not (solely) responsible for its effectiveness in resolving the paradox of certified training.

\section{Additional Discussion}
\label{app:discussion}

\subsection{Discussion on the regularization induced by IBP for large preturbations} 
In \cref{sec:deeper_nets} we show that RGS improves the performance of tight relaxations surpassing the SOTA methods on the same architecture for the setting of small perturbations. However, this improvement is not as substantial in the case of large perturbations. Based on our experimental intuitions and the findings of prior works \citep{mao2024ctbench,MaoMFV24}, we hypothesize that the regularization induced by training with IBP bounds boosts the network's certifiability in the case of large perturbations. We note that the $L_1$ regularization has been sufficiently tuned by \citet{mao2024ctbench} for MTL-IBP, thus only increasing $L_1$ regularization strength cannot achieve the kind of regularization needed for certifiability. Therefore, we speculate that large $\epsilon$ leads to much more unstable neurons, leading to exponential growth of certification difficulty. Thus, for large $\epsilon$, strong (and maybe unnecessary to robustness) regularization is required to further reduce certification difficulty. Note that the effects of this IBP regularization are more complex than just limiting the magnitude of the weights, as described by \citet{MaoMFV24}.

\subsection{Discussion on the relation between GLS and Randomized Smoothing}
\label{app:randomized_smoothing}
While Gaussian Loss Smoothing (GLS) and Randomized Smoothing (RS) \citep{CohenRK19} both involve Gaussian noise, they differ fundamentally in purpose and mechanism. RS applies smoothing in the input space, typically training on noisy inputs to enable probabilistic certification at inference time. This can be seen as a form of data augmentation and requires specialized training tailored to the target noise level.

In contrast, GLS applies smoothing in the parameter (weight) space, modifying the loss landscape to improve optimization—especially in the presence of discontinuities introduced by tight convex relaxations. As such, GLS is more closely related to classical smoothing methods used for gradient estimation and optimization in non-smooth settings.

Importantly, the certifications provided by RS are probabilistic and require substantial runtime sampling (e.g., 100+ forward passes per input), whereas our method yields deterministic certificates based on convex bounds. Although GLS-based training can be more computationally intensive during optimization, inference is straightforward and efficient, requiring no changes to the trained model.

\section{Additional Training Details}\label{app:train_details}

\subsection{Standard Certified Training}
We train with the Adam optimizer \citep{KingmaB14} with a starting learning rate of $5\times10^{-5}$ for 70 epochs on \mnist and 160 epochs on \cifar and \TIN. We use the first 20 epochs on \mnist and 80 epochs on \cifar and \TIN for $\epsilon$-annealing, with the first epoch having $\epsilon=0$ for \cifar and \TIN. We decay the learning rate by a factor of 0.2 after epochs 50 and 60 for \mnist and respectively 120 and 140 for \cifar and \TIN. For certified training on \mnist and \cifar, we use the IBP initialization proposed by \citet{ShiWZYH21}. For PGD training and for certified training on \TIN we use the Kaiming uniform initialization \citep{HeZRS15}.

\subsection{PGPE Training}
We use a training schedule of 150 epochs, with a batch size of 512 for \mnist and 128 for \cifar. We train with a starting learning rate of 0.0003 and we decay it twice by a factor of 0.4 after the 110\th and 130\th epoch. We use the first 50 epochs for $\epsilon$-annealing only when training with the large value of $\epsilon$ for each dataset (\mnist $\epsilon$ = 0.3 and \cifar $\epsilon$ = 8/255). Due to time constraints, we start all training rounds from models trained with the PGD loss in a standard gradient-based setting.

\paragraph{Training with non-differentiable bounding methods} In addition, for training with $\alpha$-CROWN + \pgpe, we use the same training schedule and hyperparameters as for standard \pgpe training. For the slope optimization procedure of $\alpha$-CROWN, we initialize all slopes with the value of 0.5 and we conduct only one optimization step with step size 0.5 for each batch, resulting in all slopes having a value of either 0.0 or 1.0. In this way, we obtain a boost in tightness when compared to standard DeepPoly, while increasing the computational cost only by a factor of 2. Slope optimization with multiple steps and smaller step sizes can further increase the tightness of the relaxation, but at the cost of increased computational complexity.

\subsection{RGS Training}

We use the same training schedules and hyperparameters as Standard Certified Training. In addition, we use a population size of $n_{ps}=2$ for all experiments, and an initial standard deviation of $\sigma_{\text{RGS}}=10^{-3}$ for all experiments. We decay the standard deviation used for sampling gradients by a factor of 0.4 at the same training steps as the learning rate. We use the same initialization schemes as for standard certified training, unless specified otherwise.

\subsection{Architectures} In \cref{tab:architectures} we present the network architectures used for all our experiments.

\begin{table}[htb]
	\centering
	\caption{Network architectures of the convolutional networks for \cifar and \mnist. All layers listed below are followed by a ReLU activation layer. The output layer is omitted. `\textsc{Conv} c h$\times$w/s/p' corresponds to a 2D convolution with c output channels, an h$\times$w kernel size, a stride of s in both dimensions and an all-around zero padding of p.}
		\begin{tabular}{cccc}
			\toprule
			CNN3-tiny & CNN3 & CNN5 & CNN5-L \\
			\midrule
			\textsc{Conv} 2 5$\times$5/2/2  & \textsc{Conv} 8 5$\times$5/2/2  & \textsc{Conv} 16 5$\times$5/2/2 & \textsc{Conv} 64 5$\times$5/2/2 \\
			\textsc{Conv} 2 4$\times$4/2/1  & \textsc{Conv} 8 4$\times$4/2/1  & \textsc{Conv} 16 4$\times$4/2/1 & \textsc{Conv} 64 4$\times$4/2/1 \\
			                             &  & \textsc{Conv} 32 4$\times$4/2/1 & \textsc{Conv} 128 4$\times$4/2/1 \\
                                         &  & \textsc{FC} 512                 & \textsc{FC} 512                 \\
			\bottomrule
		\end{tabular}
	\label{tab:architectures}
\end{table}

\subsection{Dataset and Augmentation}
\label{app:datasets}

We use the \mnist \citep{lecun2010mnist}, \cifar \citep{krizhevsky2009learning} and \TIN \citep{Le2015TinyIV} datasets for our experiments. All are open-source and freely available with unspecified license.
The data preprocessing mostly follows \citet{palma2024expressive}. For MNIST, we do not apply any preprocessing. For CIFAR-10 and \TIN, we normalize with the dataset mean and standard deviation and augment with random horizontal flips. We apply random cropping to $32 \times 32$ after applying a $2$-pixel zero padding at every margin for \cifar, and random cropping to $64\times 64$ after applying a $4$-pixel zero padding at every margin for \TIN. We train on the corresponding train set and certify on the validation set, as adopted in the literature \citep{ShiWZYH21,MullerE0V23,mao2023connecting,palma2024expressive}.

\subsection{Training costs (Time and Resources)} \label{app:train_costs}

\paragraph{Theoretical Costs of GLS} The theoretical costs of any GLS-based method scales linearly with the population size $n_{ps}$ used to obtain gradient estimates. In the case of PGPE, we need to use a large population size (e.g. $n_{ps}=256$), which makes the algorithm very costly: we need to compute the loss via forward pass $n_{ps}$ times. Estimating the gradient after computing the losses is negligible, so the total time complexity of one epoch is $O(n_{ps}*F)$, where $F$ would be the time taken by the standard certified training algorithm to only compute the forward pass in one epoch.
For RGS, there's no need to use large population sizes ($n_{ps}=2$ works well enough), so the total training time in our experiments is just $n_{ps}=2$ times larger than normal certified training (includes both forward and backward passes because we use backpropagation to obtain gradients).

\paragraph{Experimental Costs} For \pgpe and RGS training, we used between 2 and 8 NVIDIA L4-24GB or NVIDIA A100-40GB GPUs. For standard certified training and for certification of all models we used single L4 GPUs. %

In \cref{tab:train_costs} we present a detailed analysis of the training costs of the PGPE and RGS methods for all of our experimental settings (Note that the cost of \deeppoly-\pgpe for \cnnf was estimated based on training for only 1 epoch). In \cref{tab:train_costs_grad} we present the training costs for the baseline standard certified training methods for comparison.

\begin{table}[h]
    \centering
    \caption{Training costs and workload distribution across GPUs / actors for each train setting.}
    \label{tab:train_costs}
	\vspace{1mm}
    \centering
    \small
    \begin{tabular}{@{}ccllccc}
        \toprule
        Datset                 & Network          & Method & GPUs     & \makecell{Num. \\ Actors} & \makecell{Time/epoch \\(min)} & \makecell{GPU-h/ \\epoch} \\

        \midrule
		\multirow{9}{*}{\mnist} 
        & \multirow{2}{*}{\cnnty} & \deeppoly-\pgpe       & 4 x L4   & 4 & 25 & 1.73 \\
        &                         & $\alpha$CROWN-\pgpe   & 8 x L4   & 8 & 44 & 5.86 \\
        \cmidrule(rl){2-7}
        & \multirow{4}{*}{\cnnt} & \ibp-\pgpe       & 2 x L4   & 4 & 2.8 & 0.09 \\
        &                        & \crownibp-\pgpe  & 2 x L4   & 4 & 8.5 & 0.28 \\
        &                        & \hboxp-\pgpe     & 8 x L4   & 8 & 31  & 4.13 \\
        &                        & \deeppoly-\pgpe  & 8 x L4   & 8 & 27  & 3.60 \\
        \cmidrule(rl){2-7}
        & \multirow{2}{*}{\cnnf} & \deeppoly-\pgpe (est.) & 8 x L4    & 8 & $\approx300$ & $\approx40$ \\
        &                        & \deeppoly-\rgs & 8 x L4 & 8 & 7.5 & 1 \\
        \cmidrule(rl){2-7}
        & {\cnnfl} & \deeppoly-\rgs & 8 x A100 & 8 & 35 & 4.68 \\
        \midrule
        \multirow{6}{*}{\cifar} 
        & \multirow{3}{*}{\cnnt} & \ibp-\pgpe       & 2 x L4   & 4 & 6.9 & 0.23 \\
        &                       & \crownibp-\pgpe  & 4 x L4   & 8 & 8.5 & 0.57 \\
        &                       & \deeppoly-\pgpe  & 8 x L4   & 8 & 42  & 5.6 \\
        \cmidrule(rl){2-7}
        & \multirow{2}{*}{\cnnf} & \deeppoly-\pgpe (est.)  & 8 x L4    & 8 & $\approx360$ & $\approx48$ \\
        &                       & \deeppoly-RGS & 8 x L4 & 8 & 16 & 2.2 \\
        \cmidrule(rl){2-7}
        & \multirow{1}{*}{\cnnfl} & \deeppoly-RGS & 8 x A100 & 8 & 33 & 4.4 \\
        \midrule
        \multirow{1}{*}{\TIN} 
        & \multirow{1}{*}{\cnnf} & \deeppoly-\rgs & 8 x A100 & 8 & 41 & 5.5 \\
        \bottomrule
    \end{tabular}
\end{table}

\begin{table}[h]
    \centering
    \caption{Training times of \cnnf on 1xL4 GPU with standard autograd training depending on training method.}
    \label{tab:train_costs_grad}
    \begin{tabular}{clr}
    \toprule
    Dataset      & Method            & \begin{tabular}[c]{@{}c@{}}Train Time\\ (1xL4 gpu)\end{tabular} \\
    \midrule
    \multirow{5}{*}{\mnist} & PGD     & 15m           \\
                            & IBP     & 10m           \\
                            & SABR    & 20m           \\
                            & STAPS   & 25m           \\
                            & MTL-IBP & 40m           \\
    \midrule
    \multirow{5}{*}{\cifar} & PGD     & 1h30m          \\
                            & IBP     & 1h00m          \\
                            & SABR    & 2h00m          \\
                            & STAPS   & 2h30m          \\
                            & MTL-IBP & 3h10m          \\  
    \midrule
    \multirow{3}{*}{\TIN}   & PGD     & 3h15m          \\
                            & IBP     & 2h20m          \\
                            & MTL-IBP & 4h20m          \\  
    \bottomrule
    \end{tabular}    
\end{table}

\fi

\end{document}